\documentclass[mksc,nonblindrev]{informs3} 

\DoubleSpacedXI

\usepackage[draft]{hyperref}
\RequirePackage{fix-cm} 
\usepackage{xspace}
	\newcommand\ie{i.\,e.\xspace}
	\newcommand\eg{e.\,g.\xspace}

	\newcommand{\mathup}[1]{\mathrm{#1}}

	\newcommand{\e}{\mathup{e}}

  \newcommand\define{\ensuremath{\mathrel{\stackrel{\mathrm{def}}{=}}}}
  \newcommand{\abs}[1]{\left\lvert #1 \right\rvert}

\newcommand{\cmark}{\ding{51}}%
\newcommand{\xmark}{\textcolor{lightgray}{\ding{55}}}%

	\usepackage{siunitx}
    \def\sym#1{\ifmmode^{#1}\else\(^{#1}\)\fi}
    \DeclareSIUnit\eur{\officialeuro}
    \DeclareSIUnit\M{M}
    \DeclareSIUnit\k{k}
	
	\usepackage[%
  sort&compress
]{cleveref}
  \crefname{chapter}{section}{sections}
	\Crefname{chapter}{Section}{Sections}

\usepackage{etoolbox}
\robustify\bfseries

\usepackage{url}

\usepackage{isomath}
\usepackage{bm}

\usepackage{colortbl}
\usepackage{tabularx}
\usepackage{booktabs}
\usepackage{lscape}
\usepackage{collcell}
\usepackage{subfig}
\usepackage[titletoc,title]{appendix}

\usepackage{makecell}
\usepackage{float}
\usepackage{rotating}

\usepackage{rotating}

\usepackage[table]{xcolor}
\usepackage{array}
\newcolumntype{L}[1]{>{\raggedright\let\newline\\\arraybackslash\hspace{0pt}}p{#1}}
\newcolumntype{C}[1]{>{\centering\let\newline\\\arraybackslash\hspace{0pt}}p{#1}}
\newcolumntype{R}[1]{>{\raggedleft\let\newline\\\arraybackslash\hspace{0pt}}p{#1}}

\usepackage{csquotes}
\usepackage{amsmath}
\usepackage{amssymb}
\usepackage{pifont}

\usepackage{units}
\usepackage{ifthen}
\usepackage{tikz}
\usetikzlibrary{calc}
\usepackage{natbib}
\usepackage{comment}

\usepackage[
    colorinlistoftodos,
    textsize=footnotesize,
        ]{todonotes}

\makeatletter
    \renewcommand{\fps@figure}{H}         
    \renewcommand{\fps@table}{H}         
\makeatother 

\usepackage{float}
\usepackage{rotating}

\makeatletter
\newcolumntype{B}[3]{>{\boldmath\DC@{#1}{#2}{#3}}c<{\DC@end}}
\makeatother

\usepackage{natbib}
 \bibpunct[, ]{(}{)}{,}{a}{}{,}%
 \def\bibfont{\small}%

\usepackage{enumitem}

\usepackage{xcolor,colortbl}

\usepackage{tabu}
\usepackage{threeparttable}
\TheoremsNumberedThrough     

\EquationsNumberedThrough    

\def\FigureNoteName{Notes:}

\newcommand{\model}{DD-HMM\xspace}

\begin{document}



\RUNTITLE{A Duration-Dependent Latent State Model for Online Behavior}

\TITLE{Detecting User Exits from Online Behavior:\\ A Duration-Dependent Latent State Model}

\ARTICLEAUTHORS{%
\AUTHOR{Tobias Hatt}
\AFF{ETH Zurich, \EMAIL{thatt@ethz.ch}}
\AUTHOR{Stefan Feuerriegel}
\AFF{ETH Zurich \& LMU Munich, \EMAIL{feuerriegel@lmu.de}}
} 

\ABSTRACT{%
In order to steer e-commerce users towards making a purchase, marketers rely upon predictions of when users exit without purchasing. Previously, such predictions were based upon hidden Markov models~(HMMs) due to their ability of modeling latent shopping phases with different user intents. In this work, we develop a duration-dependent hidden Markov model. In contrast to traditional HMMs, it explicitly models the duration of latent states and thereby allows states to become ``sticky''. The proposed model is superior to prior HMMs in detecting user exits: out of 100 user exits without purchase, it correctly identifies an additional 18. This helps marketers in better managing the online behavior of e-commerce customers. The reason for the superior performance of our model is the duration dependence, which allows our model to recover latent states that are characterized by a distorted sense of time. We finally provide a theoretical explanation for this, which builds upon the concept of ``flow''. 
}

\KEYWORDS{hidden Markov model; latent state modeling; duration dependence; online behavior; clickstream analytics}


\maketitle
\sloppy
\raggedbottom



\section{Introduction}
\label{sec:introduction}

Most users exit e-commerce websites without making a purchase. According to a recent report, the share of users who decline to purchase accounts for more than \SI{96}{\percent} \citep{Statista.2019}. This precludes online retailers from the prospect of sales, and, hence, for marketers, there is the need of detecting users which will not purchase before they exit, so that these users can be converted into buyers. In order to detect user exits, a common approach is to model the online behavior of users and, on this basis, trigger personalized interventions \citep{Ding.2015, Montgomery.2004}.

In order to model online behavior in web sessions, marketers have previously used hidden Markov models~(HMMs) due to several strengths of this approach \citep{Ding.2015,Montgomery.2004}. First, HMMs are designed for modeling sequential data and, for that reason, have found widespread use in marketing \citep[\eg,][]{abhishek2012media,Ascarza.2013,Ascarza.2018, danaher2020tactical,Ding.2015,lemmens2012dynamics, Li.2011,kappe2018random, Montgomery.2004,Montoya.2010,Netzer.2008,park2018effects,schwartz2014model,Schweidel.2011,Zhang.2014}. In this work, we focus on web sessions that comprise a sequence of visited pages. Second, HMMs can accommodate additional information that describes the heterogeneity among both users and pages. Third, HMMs introduce latent states that can account for different phases within a web session during which users vary in their shopping behavior  \citep{Ding.2015,Montgomery.2004}.

HMMs for clickstream analytics have been previously build to understand and specifically predict user exits without purchases from web sessions. In this models, the latent states represent different user intents \citep{Ding.2015,Montgomery.2004}. User intents correspond to whether users engage in goal-directed or browsing behavior \citep{moe2003buying,Novak.2003}. The former, goal-directed behavior, refers to customers that follow a search path as they make planned purchases. For instance, this is the case when users already have an item in mind that they want to purchase. The latter, browsing, describes non-directed behavior where users engage in experiential activities. Here users scan for new shopping ideas, compare products, and thus focus largely on information collection but without making a purchase. In this paper, we connect to prior HMMs and later use them as our baselines. However, prior HMMs for clickstream analytics \citep[\eg,][]{Ding.2015,Montgomery.2004} cannot account for the duration of latent states and thus how long users have been in goal-directed or browsing behavior.

The reason why prior HMMs for clickstream analytics \citep[\eg,][]{Ding.2015,Montgomery.2004} cannot consider the  duration of latent states is that these build upon the Markov property. As such, a state can only depend on the single previous state \citep[cf.][]{Rabiner.1989}. Hence, prior HMMs can account for the duration spent on a page (\ie, how fast a user clicks) but not for the duration in a latent state (\ie, the duration of a user intent). In other words, such HMMs ignore the duration of how long a user experienced a certain latent state (\ie, how long a user experienced goal-directed or browsing behavior). While the duration spent on a page is observable and can thus be incorporated as a covariate, the duration in a latent state is unobservable and can thus only be modeled as a latent variable. As a result, the duration of a state of a ``sticky'' state may be prolonged as compared to the duration in a ``non-sticky'' state. This requires that the latent dynamics are modeled as duration-dependent. However, such duration-dependent dynamics cannot be achieved with existing HMMs for clickstream analytics and, instead, require a different model class that goes beyond the Markov property.

In this paper, we develop a \emph{duration-dependent hidden Markov model}~(\model) for predicting user exits in web sessions. For this, we extend prior HMMs for clickstream analytics by explicitly modeling transitions between latent states to depend on the duration of being in a latent state. This changes the expected duration of latent states, whereby the duration in a state is prolonged (\ie, the state becomes ``sticky''). Put simply, a greater duration of being in a certain latent state may increase the propensity to continue in this state rather than transitioning to a different state. Mathematically, we incorporate duration dependence in our model by relaxing the Markov property behind HMMs so that the latent dynamics depend not only on the single previous latent state but, rather, the duration of a latent state. 

Our empirical findings are based on actual clickstream data provided by \emph{Digitec Galaxus}, the largest online retailer in Switzerland, offering more than a million products. Our objective is to predict user exits without purchase, that is, the last page in a session that concluded without a purchase. We then confirm that our proposed \model has a superior prediction performance compared to prior HMMs for clickstream analytics: our model increases the AUROC by 7.65 percentage points. In particular, it bolsters the hit rate by \SI{18.2}{\percent}, that is, it detects an additional 18 out of 100 user exits without purchase. In particular, we find three different states that correspond to different user intents (and are named accordingly): (i)~{goal-directed}, (ii)~{``sticky'' browsing}, and (iii)~{``non-sticky'' browsing}. Here ``sticky'' (``non-sticky'') implies that users have a higher (lower) propensity to continue their online browsing activities in this latent state. We later provide a case study where we leverage the different latent states and show how our model can be used for targeting interventions in form of marketing pages with curated content.

Our research has two implications. On the one hand, the proposed model helps in better detecting users at risk of exiting without purchase. This supports marketers in managing the online behavior of e-commerce customers with the aim of increasing conversion rates, for instance, by triggering tailored interventions such as price promotions or webpage adaptations \citep[cf.][]{Ding.2015,Lam.2001,Rossi.1996,Schlosser.2006}. On the other hand, we contribute to a growing body of HMMs in marketing \citep[\eg,][]{abhishek2012media,Ascarza.2013,Ascarza.2018,danaher2020tactical,Ding.2015,kappe2018random,lemmens2012dynamics,Li.2011,Montgomery.2004,Montoya.2010,Netzer.2008,park2018effects,schwartz2014model,Schweidel.2011,Zhang.2014}. Prior HMMs in marketing rely upon the Markov property, whereas our model belongs to the more general class of hidden \emph{semi}-Markov models. Specifically, our model allows the latent dynamics to depend on the latent state duration (rather than only the latent state) and thus accommodates duration dependence. This can also be beneficial in related applications where customer behavior is subject to duration dependence (\eg, due to learning effects). 

We later explain the superior performance of our model by making a connection to the concept of ``flow''  \citep{Hoffman.1996b}. In a particular, we provide a post~hoc characterization of the latent states: (i)~Goal-directed behavior is characterized by a short duration and a comparatively high probability of purchasing. (ii)~``Sticky'' browsing primarily consists of viewing product pages that support information collection, while purchases are scarce. Due to the ``sticky'' nature of this state, it shares similarities with the concept of flow. That is, there is a prolonged latent state duration, which corresponds to a distorted sense of time. In the concept of flow, users also experience a distorted sense of time \citep{Hoffman.1996b,Hoffman.2009,Novak.2000,Novak.2003}, implying that users are in a state of deep engagement  and thus have a high propensity to remain in this state. (iii)~The state ``non-sticky'' browsing also features product pages to a large extent. However, in contrast to ``sticky'' browsing , this state is only of short duration and entails a high probability of users transitioning to a different one. Hence, this state shares similarities with non-flow behavior.

This paper is organized as follows. \Cref{sec:literature} provides background materials on clickstream analytics. In keeping with this, \Cref{sec:model} develops our \model, in which we model latent dynamics with duration dependence. We then study the properties of our model in \Cref{sec:theorymodel}. Afterwards, we introduce our empirical setup in \Cref{sec:setup}, evaluate the prediction performance  in \Cref{sec:results}, and perform a case study with marketing interventions in \Cref{sec:case_study}. Finally, \Cref{sec:discussion} discusses implications for marketers.

\section{Related Work}
\label{sec:literature} 

\subsection{User Behavior in Online Marketing}

User behavior in online marketing has been subject to extensive research \citep[\eg,][]{degeratu2000consumer, Devaraj.2002, Ghose.2012, Huang.2009, koehn2020predicting, Moe.2004, Moe.2012, Parthasarathy.1998}. One stream of literature is concerned with how user behavior is influenced by the presentation of different stimuli. In marketing, such stimuli often aim to increase conversion rates \citep{Ding.2015,Gofman.2009,McDowell.2016,park2018effects}. For instance, price promotions steer users towards increased purchasing \citep{Lam.2001,Rossi.1996}. Similarly, certain characteristics of websites, such as their formatting, their usability, or their content, are known to influence purchase decisions \citep{Ding.2015,Ludwig.2013,Palmer.2002,parboteeah2009influence,Schlosser.2006}. The effectiveness of stimuli can be further improved by personalizing them to a user's online behavior \citep{Ding.2015,Montgomery.2004}.  

A different stream of literature focuses on the navigation behavior throughout a user's web session, which is segmented according to different user intents \citep{Bucklin.2003, moe2003buying, Novak.2003, Sismeiro.2004}. User intents can be categorized by goal-oriented and browsing behavior \citep{Hoffman.1996b,Moe.2004, Nadkarni.2007}. In a goal-directed search, users make planned purchases. To this end, they adopt a utilitarian mindset in which they engage in a directed (prepurchase) search with goal-directed choices. This intent describes a cognitive state and, in order to avoid extensive cognitive effort, users conduct goal-direct search only for a short duration. In contrast, during browsing, their behavior is largely driven by information collection and is thus characterized by non-directed (ongoing) search. In this intent, users follow a experiential search path that primarily comprises navigational choices (\eg, overview pages, product pages). As a result, browsing corresponds to a hedonic mindset and thus affective thinking, which makes users perceive browsing a pleasure. 

The characteristics of user intents have implications for model development. On the on hand, user intents -- that is, goal-directed or browsing behavior -- are not constant throughout a web session but can change \citep{Hoffman.1996b,Mandel.2002}. On the other hand, user intents are usually not directly observable by marketers. Instead, they can only be recovered from other data sources such as clickstream data. Hence, this must be considered in a mathematical model by formalizing user intents as latent variables, \ie, as in HMMs \citep[see][]{Ding.2015,Montgomery.2004}.

\subsection{The Concept of {Flow} in Online Behavior}

Besides intents, the online behavior of customers is further explained by different engagement levels. This has been formalized by the so-called concept of \textquote{flow} \citep{Hoffman.1996b}. Flow refers to a state of deep engagement with a website. \citet{Hoffman.1996b} theorized flow as (i)~seamless sequences of interactions facilitated by machine interactivity, which are (ii)~intrinsically enjoyable, (iii)~accompanied by a loss of self-consciousness, and (iv)~self-reinforcing. The interaction with a website is supposed to consume a user's entire focus, such that there is little attention left to consider anything else. Due to the playful and exploratory experience, the concept of flow is especially common in online shopping \citep[\eg,][]{Koufaris.2002,MichaudTrevinal.2014,Park.2012,Tam.2006}.

Flow characterizes online behavior as follows \citep{Hoffman.1996b, Hoffman.2009,Novak.2000,Novak.2003}. When in flow, users experience a state of deep engagement, which is manifested in a distorted sense of time. As a consequence, any search activities in flow are expected to be of longer duration than activities in non-flow. As flow absorbs a user's entire attention, there is a high propensity that users will continue with activities where they remain in flow. Conversely, a non-flow state is one in which users engage comparatively less with a website as they lose interest quickly. This results in a high probability of discontinuing the current activities or even exiting the website. 

Flow is particularly applicable to users engaging in experiential search, due to the playful and exploratory nature of this experience. This was theorized earlier by \citet{Novak.2003} and has an important implication for our work: In our discussion, we later provide an explanation for the superior performance of our \model model by building upon the concept of flow. 

\subsection{Modeling Online Behavior in Web Sessions}

In order to manage customer behavior in the context of online marketing, mathematical models are needed in order to make inferences from web sessions \citep[\eg,][]{Shi.2014}. Typically, web sessions are described by clickstream data, which refers to a sequence of page visits with further heterogeneity at both the page and user levels. We refer to \citet{mobasher2007data} for an overview on clickstream analytics. The underlying objectives vary and, for instance, include predicting whether users will make purchases \citep{koehn2020predicting,Park.2016,Sismeiro.2004} or, as in this paper, to predict when users are about to exit a website without making a purchase \citep{Ding.2015,Montgomery.2004}. 

For the purpose of predicting user exits, HMMs appear beneficial, since their latent dynamics directly capture the evolution of the user intent throughout a web session \citep{Ding.2015,Montgomery.2004}. The HMM by \citet{Montgomery.2004} has served as the basis for many other works on modeling clickstream data \citep[\eg,][]{Ding.2015, hatt2020early}. The model (hereafter named ``MLSL model'') assumes that the transition probabilities between the user intent are constant over time and across users. User heterogeneity in the MLSL model is incorporated via covariates in the emission probabilities (which we also do analogously in our model). Previous works have identified two latent states that represent goal-directed and browsing behavior. Specifically, \citet{Montgomery.2004} find that users are likely to start with browsing, while a goal-directed state is attained when purchases are made. One of the states is characterized by experiential activities, while the other corresponds again to purchasing. We later use the HMM by \citet{Montgomery.2004} as our main baseline. However, prior HMMs for modeling user intents in online behavior are memory-less (\ie, duration-\underline{in}dependent). In this paper, we address this research gap by developing a tailored HMM that accommodates duration-dependent latent dynamics. 

\subsection{Hidden Markov Models in Marketing}

HMMs represent a flexible class of models with latent dynamics, whereby a sequence of observations undergoes transitions between a discrete set of unobservable (latent) states \citep{Rabiner.1989}. HMMs are based on a stochastic relationship between observations and latent state. This is particularly suitable for modeling customer behavior, where latent states reflect customer intents or other forms of engagement level \citep{Netzer.2008, park2018effects, Zhang.2014, Zhang.2016}. In the context of online behavior, latent states have been show to correspond to different user intents \citep{Ding.2015,Montgomery.2004}. In addition, latent states are often better predictors than noisy observations, thus improving prediction performance. Given these benefits, use cases of HMMs in management science are widespread \citep{netzer2017hidden}. Examples include customer relationship management \citep{Ascarza.2013,Ascarza.2018,danaher2020tactical,gopalakrishnan2021can, Kumar.2011,Luo.2013,Netzer.2008,Zhang.2014, Zhang.2016}, cross-selling analyses \citep{Li.2011,Moon.2007}, marketing resource allocations \citep{kappe2018random, Montoya.2010}, service portfolios \citep{Schweidel.2011}, and clickstream analytics \citep{Ding.2015,Montgomery.2004, Park.2016, Shi.2014}.

HMMs are based on the assumption that the latent dynamics are subject to the Markov property \citep{Rabiner.1989}. That is, latent states are only based on the single previous latent state (and not the duration of a latent state). Moreover, the duration of a latent state does not change over time and, hence, is independent of the time already spent in a state. This essentially renders HMMs \textquote{memory-less} and, as a direct consequence, prohibits a HMM from capturing engagement levels (such as those that one would expect, \eg, when experiencing a distorted sense of time), which do naturally depend on the time spent in a certain state. In contrast to this, we later develop a model in which the duration of latent states is explicitly considered, namely a duration-dependent HMM. The duration of latent states itself is latent and, for this reason, cannot be simply added to a HMM as any other observable quantity. Instead, one must turn to a different class of models, and, thus, our model falls under the wider scope of hidden \emph{semi}-Markov models \citep{Yu.2010,murphy2002hidden}.

Outside of marketing, hidden {semi}-Markov models have been previously developed \citep{Yu.2010}, even though this name leaves it unclear how the {semi}-Markov dynamics are actually modeled. One variant are variable-duration HMMs in which only the transition probabilities become duration-dependent. However, this does not change the expected duration of states. A duration-dependent HMM alters the renewal of a state, so that the distribution of durations across latent states becomes more flexible. We later develop a \model where the latent state duration follows a Weibull distribution. This differs from prior HMMs used in marketing and specifically clickstream analytics, in which, by definition, the duration is assumed to be geometrically distributed and, hence, is independent of the latent state duration. To the best of our knowledge, no previous works have tailored a duration-dependent HMM to a clickstream setting. 

In marketing, \citet{gopalakrishnan2021can} recently proposed a duration-dependent hidden Markov model for predicting customer visit frequency. However, this model is distinctly different from our \model. The main difference is that their model incorporates duration dependence in the (observable) emission component (see p.~9 in their paper), which is, in their case, whether a customer visits the shop (but not in the transition component). However, this does not allow to reveal any new and duration-dependent latent state, since, mathematically, the latent state duration can\underline{not} drive the latent variable dynamics. In contrast to this, our \model is introduces the duration dependence in the transition component, so that it can drive the latent variable dynamics. To this end, one must incorporate duration dependence into the latent state duration and, thus, into the transition probabilities. This allows us to reveal \emph{latent} user behavior that depends on the duration, as opposed to \citet{gopalakrishnan2021can} who model \emph{observed} user behavior that depends on the duration. These two approaches are, hence, orthogonal and are concerned with fundamentally different modeling questions. 

\section{The Model}
\label{sec:model} 

\subsection{Overview}

The proposed \model takes clickstream data as input in order to predict user exits without a purchase. For each web session of a user, the clickstream is given by a sequence of visited pages. Mathematically, we refer to pages by an observable variable $O_{it} \in \mathcal{O}$ for a given user $i$ at time period $t \in \mathbb{N}_{\geq 0}$. If a user exits a web session without a purchase, we denote the last page of her sequence by $O_{it} \define \textsc{Exit}$. Based on this, the objective is to predict the page at which a user exits without purchase, that is, estimating $P(O_{it} = \textsc{Exit}\mid O_{i1}, \ldots, O_{i,t-1})$. 

Our model accounts for different user intents. These are assumed to be not directly observable and, instead, they are only stochastically related to the observable page sequence. Mathematically, we capture different phases in online behavior by modeling them as latent variables. Later, we find three different states and, consistent with \citep{Ding.2015,Montgomery.2004}, label them as: (i)~{goal-directed}, (ii)~{``sticky'' browsing}, and (iii)~{``non-sticky'' browsing}. In our discussion, we later also provide an explanation by linking the latent states to the concept of flow. 

We need a tailored model to capture duration dependence. For this purpose, simply modeling the evolution of latent states with a Markov chain is not sufficient. This would yield in an existing HMM for clickstream analytics \citep[\eg,][]{Montgomery.2004} where the next latent state is based only on the previous latent state. Rather, the latent dynamics must consider the duration of a latent state. For this purpose, we extend prior HMMs for clickstream analytics by modeling not only the latent states but also the latent state durations.

Importantly, both the time spent on a page and the duration in a latent state are different concepts. The former measures how long a user has viewed a specific page before continuing (\eg, a user might have looked at a product page for 3 seconds). The latter, the duration in a latent state, refers to for how many pages a users has experienced a specific user intent. For example, a user might has been in goal-directed search for the last three page visits. In this regard, it is further crucial to emphasize that (i)~the time spent on a page as it is considered in \citet{Montgomery.2004} and (ii)~the duration in a latent state correspond to vastly different behavior. The former considers the duration of a click; the latter considers the duration of a latent state and thus of a user intent. Put simply, the former considers how long users viewed a page, the latter how long they have been in goal-directed or browsing behavior. This has also direct implications for modeling: the former is based on an observable quantity, which can be easily added in the model, while the latter is an unobservable quantity, which must be modeled as part of the latent dynamics.\footnote{\SingleSpacedXI\footnotesize One might simply think that one could adapt the HMM of \citet{Montgomery.2004} by changing the diagonal elements in the transition matrix and yield the same behavior. However, this is not the case. On the contrary, simply changing the diagonal elements in the transition matrix does not change the distribution of the latent state duration. It would still remain geometrically distributed and, as a consequence, could not capture states that become \textquote{stickier} over time. In addition, transition probabilities would remain constant over time and, in particular, would not be dependent on the latent state duration.} We elaborate in depth on differences between the HMM from \citet{Montgomery.2004} (called MLSL model) and our \model in \Cref{sec:comparison_MLSL}.

Duration dependence has been previously studied in marketing research but, generally, outside of the context of HMMs. For instance, a customer's propensity to churn is not constant in time but dependent on the duration of being a customer  \citep{Fader.2018}. In this work, we introduce a duration-dependent HMM. We later show that prior HMMs are a special case of our \model under specific conditions. The extension from prior HMMs to a duration-dependent HMM has profound implications: the durations of latent states are also latent and, owing to this, we cannot simply insert them as covariates to the transition mechanism.

Our model distinguishes (i)~duration dependence in transitions where the previous state is maintained (\ie, self-transitions) and (ii)~duration dependence in transitions where the model switches to a different state. The former, duration dependence in self-transitions, affects the expected duration of latent states. Hence, the model can prolong the duration of a ``sticky'' state as compared to a ``non-sticky'' state. Mathematically, this is achieved through flexible distributions of latent state durations. The latter, transitions between different states, changes the propensity of new states depending on the current engagement level (\eg, longer duration in a state with deep engagement makes it more likely that the next state will be one of low engagement). Mathematically, this is modeled by duration-dependent transition probabilities.   

\subsection{Model Specification}

\begin{figure}[h!]
	\SingleSpacedXI
	\FIGURE
	{\includegraphics[scale=0.5]{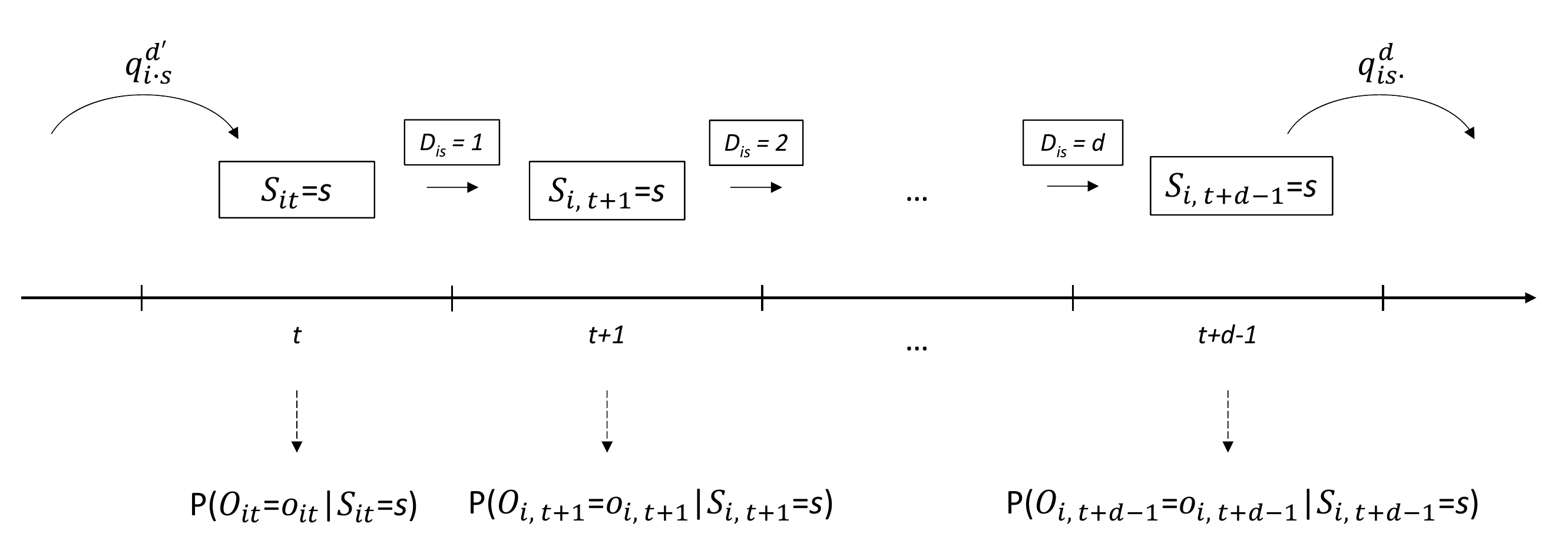}}
	{Duration-Dependent Hidden Markov Model.\label{fig:model_overview}}
	{At time period $t$, the user attains a latent state $s$ (\ie, an intent), which is associated with a certain duration $D_{is}$. Hence, unlike prior HMMs for clickstreams, the latent states does not change during the subsequent periods, but remains the same for a duration of $D_{is} = d$ time periods. Afterwards, the latent state changes to a different one ($s' \neq s$). Here the transition probability is dependent on the duration of the current latent state.}
\end{figure}

The proposed \model consists of four components (see \Cref{fig:model_overview}): (i)~the emission component, specifying the distribution of the pages visited in each state; (ii)~the initial latent state distribution; (iii)~the distribution of latent state durations; and (iv)~the duration-dependent transition mechanism. Here the latter two components are responsible for the duration dependence, which constitutes the difference from prior HMMs. 

\subsubsection{Emission Component.}

The emission component links the observable quantities to the latent states. We denote the latent states by $S_{it} \in \mathcal{S}$ where $K = \abs{\mathcal{S}}$ is the number of different latent states. We then model the observed page $O_{it}$ as a stochastic realizations of a latent state $S_{it}$, which a user $i$ occupies during the time period $t \in \mathbb{N}_{\geq 0}$. Here $t$ refers to the $t$-th page in a web session. Formally, the emission $p_{it\mid s}^o$ describes the probability of visiting a page $o \in \mathcal{O}$ during time period $t$ depending on a user's current latent state $s$. The emission is given by
\begin{equation}
p_{it\mid s}^o = P\left(O_{it} = o \mid S_{it} = s\right). 
\end{equation}

Analogous to \citet{Netzer.2008}, the probability $p_{it\mid s}^o$ is modeled as a logit model. In our work, this allows us to accommodate further information on heterogeneity at both page and user level. Hence, the probability of observing certain pages is influenced by an additional set of covariates $x_{it}$. The covariates can potentially be time-varying. For instance, common choices of such covariates \citep{Ding.2015, Zhang.2014} are the time spent on the last page or the depth of the current session until $t$ (\ie, the number of previously visited pages). We later also use other covariates such as, \eg, a weekend dummy and customer type. Here one can also include covariates on marketing interventions (if such interventions were previously observed). We use the same covariates for the emission probabilities as in \citet{Montgomery.2004}. 

Note that these covariates are different from the duration of the latent state: covariates such as the time spent on a page and the depth of a session can be observed. Yet, the duration of the latent state is itself latent and, hence, cannot be observed. Because of this, the duration in a latent state cannot be included in the covariates but must rather be modeled as a separate latent variable. Formally, the logit model is given by
\begin{equation}\label{eq:emission}
	p_{it\mid s}^o = \frac{\e^{\gamma_{is}^o + \beta_{is}^o x_{it}}}{\sum\limits_{k \in \mathcal{O}} \e^{\gamma_{is}^k + \beta_{is}^k x_{it}}} 
\end{equation}
with intercept $\gamma_{is}^o$ and coefficients $\beta_{is}^o$. The intercept captures the general propensity towards visiting certain pages, while the latter quantifies the (time-varying) influence of the covariates. Both parameters, intercept $\gamma_{is}^o$ and coefficients $\beta_{is}^o$, are later extended by a hierarchical structure in order to accommodate further information on between-user heterogeneity.

\subsubsection{Initial Latent State Distribution.}

The initial latent state distribution describes the probability of a user starting the session in latent state $s$. It is given by $\pi = [\pi_{1}, \pi_{2}, \dots,  \pi_{K}]$ for all $K$ states, where $\pi_{s} = P\left(S_{i1} = s\right)$ for $s$ in $\mathcal{S}$ and $\sum_{i=1}^K\pi_{i} = 1$. The initial state distribution is later estimated from the clickstream data.

\subsubsection{Distribution of Latent State Durations.}
\label{sec:distribution_latent_state_durations}

In our \model, we explicitly model the duration $D_{is} \in \mathbb{N}_{\geq 0}$ for each latent state. Specifically, our \model allows for a flexible state-specific duration distribution. This is different from prior HMMs, where the duration of each state is geometrically distributed. Hence, in prior HMMs, a new state can be attained after each time period (\ie, $D_{is} = 1$ for all $s$). In contrast, the durations in our \model vary, and, on top of that, states might occur for more than one time period (\ie, arbitrary, state-specific durations $D_{is} \in \mathbb{N}_{\geq 0}$). As a result, our \model can prolong the duration of certain states such that the underlying states become ``sticky''.

Let $D_{is}$ denote the duration of user $i$ being in state $s$. In consumer research, this duration is sometimes also referred to as sojourn time. We model the duration via a discrete Weibull distribution.\footnote{\SingleSpacedXI\footnotesize By nature, the Weibull distribution is continuous in time. However, the duration $D_{is}$ of a state $s$ is discrete in time, since we consider time periods. We thus use the discretized variant introduced by \citet{Nakagawa.1975}.} The Weibull distribution has previously been used in marketing literature \citep[\eg,][]{Fader.2018} for modeling duration dependence, albeit outside of a latent state model. In our \model, the duration is then modeled via
\begin{equation}\label{eq:weibullsojourn}
	P\left(D_{is} = d\mid \theta_s, c_s\right) = (1-\theta_s)^{(d-1)^{c_s}} - (1-\theta_s)^{d^{c_s}} \quad \text{ for }0<\theta_s<1,\ c_s>0,
\end{equation}
with parameters $\theta_s$ and $c_s$. Both quantify the overall \textquote{stickiness} of latent states, where $\theta_s$ represents the overall magnitude of the \textquote{stickiness} and $c_s$ its prolongation. \Cref{eq:weibullsojourn} describes the distribution of the latent states. As a result, some states (\eg, when a user is in a state of deep engagement) can occur for a longer duration than others. This is controlled by the parameters $\theta_s$ and $c_s$. In order to understand how these parameters influence the model behavior, we later perform an analysis with simulated data (see \Cref{sec:theorymodel}).  

\subsubsection{Duration-Dependent Transition Mechanism.}

The transition mechanism is responsible for modeling the evolution of latent states. In the context of our work, latent states can change during a web sessions as users collect more information. Hence, there is a certain probability that a user transitions to a different latent state. In our \model, the transition mechanism is designed in such a way that it accommodates duration dependence. Hence, the transition probability is dependent on the previous duration of a user being in a latent state.

In the following, we model transitions among \emph{different} states ($s' \neq s$). We explicitly set the transition probability for transitions within the same state (self-transitions) to zero. This is done because self-transitions are already considered as part of the latent state durations; see \Cref{sec:distribution_latent_state_durations}.

Formally, the duration-dependent transition matrix is defined as
\begin{equation}
\SingleSpacedXI
Q_i^d = \begin{pmatrix} 
    0 & q_{i12}^d & \dots  &q_{i1K}^d\\
	q_{i21}^d & 0 & \dots  &q_{i2K}^d\\
    \vdots & \vdots&\ddots  &\vdots\\
    q_{iK1}^d & q_{iK2}^d &\dots  &   0 
    \end{pmatrix},
\end{equation}
where $q_{iss'}^d$ denotes the probability that user $i$ changes to state $s'$ given the current state $s$ and its duration $d$. The diagonal elements are constrained to be zero, since self-transitions are part of the latent state durations. Note that the transition matrix $Q_i^d$ depends on the duration $d$. This allows some states to become less relevant compared to others over time by reducing the transition probability as a function of $d$. The actual transition probabilities $q_{iss'}^d$ are modeled via a logit model 
\begin{equation}\label{eq:transitionprob}
	q_{iss'}^d = \frac{e^{\mu_{iss'} + \delta_{iss'}d}}{\sum_{k\in\mathcal{S}\setminus \{s\}}{e^{\mu_{isk} + \delta_{isk}d}}},
\end{equation}
where the parameters $\mu_{ss'}$ determine a user's propensity to transition from state $s$ to $s'$ and where $\delta_{ss'}$ captures the effect of the state duration $d\in \mathbb{N}_{\geq 0}$ on the transition probability. \Cref{eq:transitionprob} also highlights why extending an HMM towards duration dependence is non-trivial: the transition mechanism, which already models the latent dynamics, depends now on an additional latent quantity, \ie, the latent state duration $d$.

The probability of transitioning from state $s$ to state $s'$ depends solely on $\mu_{ss'}$ (the user's propensity to transition from state $s$ to state $s'$) and $\delta_{ss'}$ (which controls the dependence of the transition probability on the state duration).\footnote{\SingleSpacedXI\footnotesize Note that the parameter $\delta_{ss'}$ is defined per state, \ie, our \model can capture scenarios where transitions are duration-independent in some states, but are duration-dependent in others.} If $\delta_{ss'}$ equals zero, the transition probability from state $s$ to state $s'$ remains constant irrespective of the duration of state $s$. If $\delta_{ss'}$ differs from zero, the state duration either increases or decreases the probability of transitioning from the current state to another. 

We also estimate a version of our \model which accounts for unobserved heterogeneity in the transition probabilities. We achieve this by allowing the threshold parameters $\mu_{ss'}$ to vary across individuals. However, we find that this leads to overfitting the data. Details are provided in \Cref{sec:results} and \Cref{appx:sec:DDHMM_UH}.

\subsection{Comparison to the HMM in \citet{Montgomery.2004} (``MLSL Model'')}
\label{sec:comparison_MLSL}

In this section, we compare our \model to a common HMM used to model online behavior based on clickstream data, namely the HMM from \citet{Montgomery.2004}. We refer to the HMM from \citet{Montgomery.2004} as MLSL model. The MLSL has served as the basis for many other works on modeling clickstream data \citep[\eg,][]{Ding.2015, hatt2020early}. 

The MLSL model assumes homogeneous transition probabilities (\ie, homogeneous across time, which enforces the Markov property) and incorporates user heterogeneity in the emission probabilities via covariates similar to our model. Our \model extends the MLSL model by relaxing the Markov property. The resulting model belongs to the wider class of hidden semi-Markov models \citep{Yu.2010}. In our work, we extend the MLSL model in two directions: (i)~We allow for flexible state duration distributions and (ii)~capture transition probabilities that are duration-dependent. Both are discussed in the following.

The MLSL model (because of its Markov property) has a latent state duration that is geometrically distributed. This has direct implications: the renewal probability (\ie, the probability of self-transitions where by HMM remains in the current state) is constant in time.\footnote{A derivation of this fact is provided in Appendix~\ref{appendix:distribution_latent_state_durations}.} As such, the probability of remaining in a latent state does not depend on the duration of the state. Hence, the MLSL model cannot accommodate states that are ``sticky''. However, the latter is desirable to represent, \eg, a distorted sense of time or states of deep engagement. In contrast, our \model encompasses a general state duration distribution. Here we use a Weibull distribution, so the renewal probability of remaining in the state is duration-dependent and, as a result, we yield latent state durations of different lengths. Notwithstanding, our model can also recover latent states found in \citep{Ding.2015,Montgomery.2004}, since it allows the renewal probability to be constant in time. This is due to the fact that, for $c_s=1$, the Weibull distribution reduces to a geometric distribution.

The transition probabilities in the MLSL model are constants. That is, the duration of the current latent state does not influence the transition probability to another state. Hence, the MLSL model (and any variant based on it) cannot capture any information that goes beyond the current state and, as such, is memory-less. Our \model explicitly allows the probability of transitioning from the current state to another state to depend not only on the current state, but also on the time elapsed in this state (see \Cref{eq:transitionprob} for details). This component introduces a memory property as the transitions depend on the current state \emph{and} the duration of this state. Similar to before, our \model is still able to recover the same latent states as in \citet{Ding.2015} and \citet{Montgomery.2004}: for $\delta_{ss'}=0$, our \model is able to accommodate memory-less transition probabilities without duration dependence as in \citep{Ding.2015,Montgomery.2004}.

\subsection{Model Estimation}
\label{sec:model_estimation}

We combine the parameters of the \model model in the set $\lambda_i = \{(Q_i^d)_{d\geq 1}, p_{it\mid s}^{o}, \theta, c,\pi\}$, where $(Q_i^d)_{d\geq 1}$ describes the duration-dependent transition probabilities, $p_{it\mid s}^{o}$ denotes the emission probabilities for each state and each page, $\theta$ and $c$ determine the distribution of the state duration, and $\pi$ is the initial state distribution.

The above model is estimated using a hierarchical Bayesian framework  \citep{Rossi.1996}. This allows us to capture additional information on the heterogeneity across pages and users. Specifically, we accommodate additional covariates in the emission component -- that is, intercept $\gamma_{is}$ and coefficients $\beta_{is}^{o}$ -- as follows. We assume that both parameters have random coefficient specifications and further follow a multivariate regression, \ie,
\begin{equation}
		\gamma_{is} = \Pi_{s} R_{i} 
		\qquad\qquad\text{and}\qquad\qquad
		\beta_{is}^{o} = \Phi_{s}^o R_{i},
\end{equation}
where $R_i$ is the vector of demographic measures (including an intercept) for user $i$ and where matrices $\Pi_{s}$ and $\Phi_{s}^o$ contain parameters similar to \citet{Montgomery.2004}. Randomness is induced by choosing an appropriate prior distribution over the parameter matrices $\Pi_{s}$ and $\Phi_{s}^o$.

We estimate all model parameters in the \model using a Bayesian approach.\footnote{\SingleSpacedXI\footnotesize We ran additional robustness checks concerning our model specification in order to ensure that we can retrieve the original parameters from simulated data. All checks turned out positive. See \Cref{apx:simulation_study} for details.} Specifically, we use Markov chain Monte Carlo (MCMC) sampling. The number of latent states, $K$, is determined by comparing model fits across different parameter values. Details on our estimation procedure are provided in Appendix~\ref{appendix:estimation_details}.

\section{Model Properties}

\label{sec:theorymodel}

In the following, we study how different parameters of the \model affect its behavior, with respect to (i)~state durations and (ii)~renewal probabilities. The latter is introduced later as a derived quantity in order to facilitate better interpretability.

\subsection{State Duration}

\begin{figure}[h!]
\SingleSpacedXI
    \centering
  	\caption{Example State Duration Across Different Values of $\theta_s$ and $c_s$ for the Discrete Weibull Distribution as a Function of the Latent State Duration $d$ (\ie, Number of Pages).}
    \input{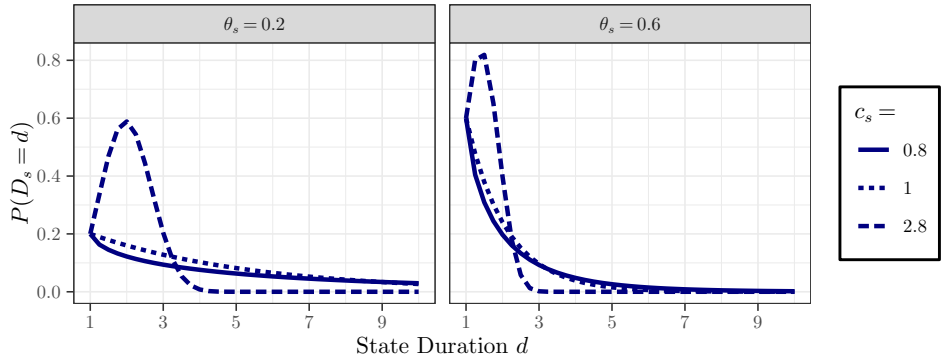}
	\label{fig:theoweilbulldist}
\end{figure}

The state duration in our \model follows a discrete Weibull distribution. \Cref{fig:theoweilbulldist} shows the expected state duration for different parameters $\theta_s$ and $c_s$. The former, $\theta_s$, quantifies to what extent a state is \textquote{sticky}. This scales the curves along the $y$-axis. The latter, $c_s$, refers to the state's prolongation on the time dimension. Hence, the latter is of particular importance for changing the overall duration of states (\ie, $D_{is})$. 
\begin{quote}\begin{quote}%
		\begin{itemize} 
			\item For $c_s  > 1$, the state is \textquote{sticky} for only a short duration. This is illustrated in the figure, where we observe a rapid increase followed by steep decline and a probability that approaches zero. 
			\item For $c_s < 1$, the \textquote{stickiness} of a state is prolonged. In particular, the likelihood of long latent state durations is increased and remains positive. This is illustrated in the figure where the initial spike is less rapid but where the overall decline is much lower. 
			\item For the case $c_s=1$, the discrete Weibull distribution reduces to the geometric distribution (as in, \eg, \citet{Ding.2015} and \citet{Montgomery.2004}). Hence, the state duration becomes the same as in traditional HMMs. Specifically, the expected latent state duration then computes to $\mathbb{E}[D_{is}] = 1$ (see Appendix~\ref{appendix:distribution_latent_state_durations}). 
		\end{itemize}
\end{quote}\end{quote}

The Weibull distributed latent duration is a distinct feature of our \model that sets it apart from prior HMMs for modeling online behavior. The reason for this is that the Weibull distribution allows the \textquote{stickiness} to change over time leading to an increasingly deeper user engagement. 

\subsection{Renewal Probability}

Let us now assume that a user is in a state of deep engagement, which implies that (s)he is likely to remain in that state and thus introduces a ``stickiness'. However, when not in a state of deep engagement, a user is likely to be bored and thus more included to change to a different state.

To better understand how such dynamics can be captured by our model, we introduce an additional quantity, the renewable probability $\rho_s(d)$. It represents the probability of remaining in (\ie, renewing) the current state $s$ after $d$ periods have already elapsed in this state. Formally, it is defined by
\begin{equation}\label{eq:geomprobremaining}
	\rho_s(d) = P\left(D_{is} \geq d+1 \mid D_{is} \geq d\right).
\end{equation}
We can rewrite $\rho_s(d)$ for when the state duration follows a Weibull distribution. We then yield 
\begin{equation}\label{eq:dwprobremaining}
	\rho_s(d\mid\theta_s, c_s) = P\left(D_{is} \geq d+1\mid D_{is} \geq d, \theta_s, c_s\right) = \frac{(1-\theta_s)^{(d+1)^{c_s}}}{(1-\theta_s)^{d^{c_s}}} = (1-\theta_s)^{(d+1)^{c_s}-d^{c_s}}.
\end{equation}
As can be seen from the term $(d+1)^{c_s}-d^{c_s}$, the renewal probability is subject to duration dependence. Here the parameter $c_s$ is again of particular relevance, as it governs the behavior of the renewal probability. This can be seen by comparing different cases for the parameter $c_s$. If $c_s > 1$ ($c_s < 1$), the probability of renewing a state declines (increases) and the state becomes less (more) \textquote{sticky}. In the special case of $c_s = 1$, the distribution reduces to a geometric distribution and thus yields a traditional HMM where the latent dynamics are duration-independent.

\begin{figure}[h!]
\SingleSpacedXI
    \centering
	\caption{Example Renewal Probabilities as a Function of the Latent State Duration $d$ (\ie, Number of Pages).}
    \input{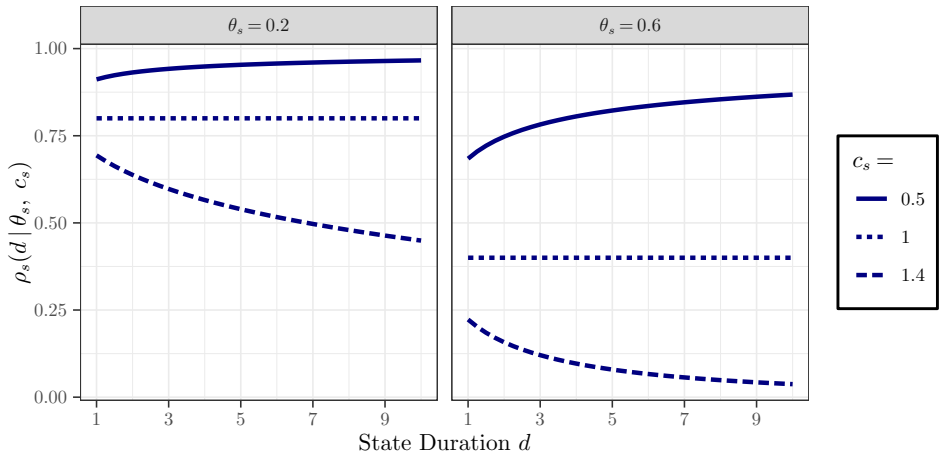}
	\label{fig:theorenewalprob}
\end{figure}

The renewal probability of a state $\rho_s(d\mid\theta_s, c_s)$ given the discrete Weibull distribution is a function of two parameters, namely $\theta_s$ and $c_s$. The parameter $\theta_s$ models the overall magnitude of the \textquote{stickiness}, whereas the parameter $c_s$ allows to add a temporal structure to the otherwise constant renewal probability. This introduces the following characteristics (see \Cref{fig:theorenewalprob}):
\begin{quote}\begin{quote}%
\begin{itemize} 
\item For $c_s>1$, the term $(1-\theta_s)^{(d+1)^{c_s}-d^{c_s}}$ from \Cref{eq:dwprobremaining} increases in the state duration $d$. This means that the longer the duration of a state, the less likely a user is to renew that state. In this case, the discrete Weibull distribution exhibits negative duration dependence. 
\item In contrast, for $c_s<1$, the term $(1-\theta_s)^{(d+1)^{c_s}-d^{c_s}}$ decreases in the state duration $d$. This means that the longer the duration of a state, the more likely a user is to renew that state. We refer to this as positive duration dependence. 
\item For $c_s = 1$, the renewal probability of a state remains constant for a geometrically distributed state duration. For $c_s=1$ and duration-independent transitions, we obtain the dynamics as in prior HMMs \cite[\eg,][]{Montgomery.2004, Ding.2015} as a special case. 
\end{itemize}
\end{quote}\end{quote}
\noindent
One might think that traditional HMMs such as the one in \citet{Montgomery.2004} can also capture \textquote{stickiness} by a high probability of self-transition. However, this is not the case. As we have seen above, the traditional HMM has geometrically distributed latent durations (\ie, $c_s=1$). Hence, using \Cref{eq:dwprobremaining} with $c_s=1$, we see that $\rho_s(d\mid\theta_s, c_s) = 1-\theta_s$. As a consequence, the renewal probability is duration independent and, hence, does not change over time. Put simply, the \textquote{stickiness} does not change over time. As such, the traditional HMM cannot capture user engagement that becomes increasingly more ``sticky''. Opposite to this, our \model is able to capture such dynamics: the Weibull distribution allows for a renewal probability that changes over time and, hence, allows the \textquote{stickiness} to change over time. Since our \model incorporates a Weibull-distributed latent duration, it is able to capture a changing \textquote{stickiness}.

\section{Empirical Setup}
\label{sec:setup}

\subsection{Data Collection}

Our evaluation is based on real-world clickstream data provided by our partner company, \textit{Digitec Galaxus}. Our partner company is the largest online retailer in Switzerland and offers more than a million different products, including clothing and, to a larger extent, electronic devices. Its website provides additional information that should facilitate the search process of potential customers, such as product reviews, overview pages, and curated marketing pages. As with other online retailers, the predominant share of users exits their web session without making a purchase. For reasons of confidentiality, we cannot disclose the exact conversion rate but note that it is substantially above those of other online retailers for whom exits without purchases have been estimated to occur in 96\,\% of all web sessions \citep{Statista.2019}. Hence, there is a clear need to identify users at risk of exiting without a purchase so that potential interventions for managing conversion rates can be triggered.

We intentionally design our model to be applicable to cold-start settings \citep[\eg,][]{padilla2021overcoming}, \ie, without the use of customer variables. The reason for this is that our partner company informed us that most users are not ``logged in'', and, hence, customer variables cannot be obtained for them. In fact, this applies to more than 90\,\% of the users. This modeling decision is also due to European regulation promoting privacy (\eg, EU General Data Protection Regulation). We later discuss how our model can be extended to accommodate customer variables as part of the covariates (see \Cref{sec:discussion}). To this end, we explicitly control for unobserved heterogeneity through our random coefficient specification of our Bayesian hierarchical approach.

Our data consists of 400 web sessions from July 2019.\footnote{\SingleSpacedXI\footnotesize For reasons of confidentiality, we cannot disclose the exact time period as it would disclose the overall web traffic to competitors.} The data collection was conducted analogously to earlier research \citep{Montgomery.2004}: (i)~Sessions from web crawlers or other bots that did not originate from humans were discarded (using a company internal software). (ii)~We excluded sessions with fewer than three and more than 50 page visits.

\subsection{Data Preprocessing}

Similar to \citet{Montgomery.2004}, we annotated pages according to the underlying page categories. This yields the observed page sequence $O_{it} \in \mathcal{O}$ with $\mathcal{O} = \{$\textsc{Home}, \textsc{Account}, \textsc{Overview}, \textsc{Product}, \textsc{Marketing Page}, \textsc{Community}, \textsc{Checkout}, \textsc{Order}, \textsc{Exit}$\}$. It is possible that the user purchases a single or even multiple items within the same session. However, if this is not the case, we labeled the last page in a session without purchase as $O_{it} \define \textsc{Exit}$. Marketing pages are curated pages with additional information on specific product types, thereby guiding customers in the purchase funnel. Our partner company uses such marketing pages in order to set themselves apart from other market players and, in particular, from new market entries. We later examine the role of such marketing pages in purchasing as part of our case study (see \Cref{sec:case_study}). In the following, our objective is to predict the latter label, that is, user exits without purchase as given by $O_{it} = \textsc{Exit}$. Throughout our subsequent analyses, the objective is to achieve high prediction performance in this task.

We include additional covariates $x_{it}$ that capture heterogeneity at both page and user level. This choice is informed by prior research \citep{Ding.2015,Montgomery.2004}. We computed: (i)~the visit depth, \ie, the number of pages visited\footnote{\SingleSpacedXI\footnotesize The identification between visit depth and latent state duration is ensured. The reason is that user intents are known to change during web sessions \citep{Hoffman.1996b,Mandel.2002}, and, hence, users typically attain more than one latent state per session. Hence, both variables, visit depth and latent state duration, are barely correlated. Nevertheless, we performed a robustness check in which we assessed whether we can retrieve the original parameters from simulated data. This check led to a positive result.} until time period $t$ during the sessions; (ii)~the time spent on each page (in seconds); (iii)~the cumulative number of visits to the same page (\ie, when certain pages are visited repeatedly); (iv)~a weekend dummy ($=1$ if weekend); and (v)~proxies for a variety of customer variables (such as customer type, \ie, whether the customer is a private person or a company), which our partner company infers from the IP address through a proprietary algorithm.

The covariates such as visit depth and duration are (re)computed dynamically for each user $i$ and time period $t$. Hence, they are (re)computed at every time period. As such, all covariates are available at every time period in the session and can be used for prediction at every time period in the session. This is important, since a prediction need to be available at any time period in order to steer the user during the session. In particular, the latent state, which is used to recommend different interventions can be computed already at the first page visited (using the initial state probabilities). The complete list of covariates is reported in \Cref{tbl:SummaryStatistics}.

Importantly, dynamic interventions (\eg, coupons, price promotions) are currently not implemented at our partner company due to its positioning in a high-priced customer segment. As such, we cannot enter corresponding data on dynamic interventions into our model. However, if such interventions were available, marketers can update our model straightforwardly, simply by including the interventions as further covariates into our model.

All covariates are dynamically (re)computed with data with data until time period $t$ (and, thus, without lookahead bias). In particular, covariates such as visit depth and duration are recomputed at every time period. As such, all covariates are available at every time period in the session and can be used for prediction at every time period in the session. This is important, since a prediction need to be available at any time period in order to steer the user during the session. In particular, the latent state, which is used to recommend different interventions can be computed already at the first page visited (using the initial state probabilities). 

We split the data into a training set (75\,\%) and a test set (25\,\%). Our split accounts for both the user and the time level. Specifically, we make the split at the level of user sessions, so that we use one set of user sessions exclusively for training, while a separate set of user sessions gives the test set for measuring the out-of-sample performance. Moreover, we ensure that a correct order in time is maintained. For this, we make sure that all user sessions in the test set take place after the sessions in the training set. Later, the user sessions in the test set are used for reporting the out-of-sample prediction performance (that is, on unseen observations). 

\subsection{Summary Statistics}
\label{sec:sumstatistics} 

\Cref{tbl:SummaryStatistics} reports summary statistics for the clickstream data. On average, a session lasted for 6.04 minutes (standard deviation: 9.31) and consisted of 9.80 pages (standard deviation: 6.33). Sessions which led to a purchase lasted considerably longer in terms of time (17.10 vs. 2.16 minutes), but not in terms of the number of visited pages (10.09 vs. 9.71 pages per session).  Overall, users spent most of the session on \textsc{Product} pages, follow by \textsc{Overview} pages. In addition, summary statistics for the covariates are provided in \Cref{tbl:SummaryStatisticsCovariates}.

\begin{table}
\SingleSpacedXI
\footnotesize
\centering
\caption[]{Summary Statistics across of Clickstream Data. \label{tbl:SummaryStatistics}}
\begin{tabular}{l SS SS SS}  
\toprule
&\multicolumn{2}{c}{Overall}& \multicolumn{2}{c}{Exit}& \multicolumn{2}{c}{Purchase} \\
\cmidrule(r){2-3}
\cmidrule(r){4-5}
\cmidrule(r){6-7}
& {Mean} & {Std. dev.} & {Mean} & {Std. dev.}& {Mean} & {Std. dev.}\\
\midrule
\csname @@input\endcsname Summary_statistics_digi 
\bottomrule
\multicolumn{7}{l}{\footnotesize{$^a$\textsc{Order} and \textsc{Exit} pages are not included for reasons of confidentiality.}}
\end{tabular}
\end{table}

\begin{table}
\SingleSpacedXI
\footnotesize
\centering
\caption{Summary Statistics of Covariates $x_{it}$.\label{tbl:SummaryStatisticsCovariates}}
\begin{tabular}{l SSSSS}  
\toprule
& {Mean} & {Std. dev.} & {Min} & {Median}& {Max}\\
\midrule
\csname @@input\endcsname Covariates_table 
\bottomrule
\multicolumn{6}{p{9.25cm}}{\emph{Note:} All covariates are dynamically (re)computed at each time period $t$ (without lookahead bias).}
\end{tabular}
\end{table}

\subsection{Performance Metrics}

The performance in terms of predicting user exits without a purchase is measured on unseen data, that is, on the test set. For this, we draw upon the following metrics:
\begin{quote}\begin{quote}
\begin{itemize}
\item \textbf{AUROC:} The area under the receiver operating characteristic curve (AUROC) quantifies the trade-off between the false positive rate and the true positive rate. Formally, it refers to the performance of detecting users that have exited without a purchase relative to how many users were falsely detected across all classification thresholds. An AUROC of 0.5 corresponds to the performance of a majority vote, while 1 represents a perfect classifier. 
\item \textbf{AUPRC:} We also compare the area under the precision recall curve (AUPRC). The AUPRC is often preferred for problems with imbalanced data where one class (here: exits without a purchase) is of particular interest \citep{Davis.2006, Saito.2015, Sofaer.2019}. 

\item \textbf{Hit rate:} The hit ratio provides the ratio of correctly identified user exits without a purchase among all sessions. The hit ratio focuses on the performance that is achieved for the positive class (\ie, exits without a purchase), as this class is of interest to marketers. The hit rate depends on the choice of threshold, which was provided by our partner company, since the threshold determines the amount of marketing interventions (such as price promotions) spent. We report the hit rate at a false positive rate of \SI{30}{\%}. A relatively low false positive rate is chosen in order to avoid too many false positive alarms and, therefore, erroneously implemented online interventions (\eg, price promotions).
\end{itemize}
\end{quote}\end{quote}

\subsection{Benchmark Models}

We compare our \model against several baselines from the literature as follows. Note that all baselines have access to the identical set of covariates as our \model. Moreover, we repeat that all features are (re)computed dynamically for each user $i$ and each time period $t$, but only with information available until that time period. The baselines are:
\begin{enumerate}
\item Logistic regression as in \citep{Sismeiro.2004}. 
\item Random forest to handle non-linear relationships \citep[cf.][]{breiman2001random}. 
\item First-order Markov model, which estimates the transition probabilities between the page categories. 
\item Long-short term memory (LSTM) network \citep{koehn2020predicting}. The LSTM is sequential neural network \citep{kraus2020deep}, which possesses several similarities with HMMs. In particular, the LSTM also extracts an embedding (in form of a latent variable) that represents the sequence of pages visited. However, different to HMMs, the latent variables in the LSTM model are not interpretable, since the take values in a high-dimensional vector space. As such, the latent variable cannot be (interpretably) connected to certain user behavior, which is possible in the HMM due to its discrete latent variable. As such, LSTMs are not helpful for supporting marketing interventions in online environments. The reason being is that the latent states capture user intents and must thus be interpreted by marketers such that interventions can be tailored based on the latent user intent. 
\item Markov modulated marked point process model~(M3PP) from \citet{hatt2020early}. This model also captures the users' latent states, but, different from this work, is tailored to the continuous-time setting. 
\item Static HMM as a baseline in which transitions between states are prohibited. The static HMM also comprises different latent states; however, transitions between them are disabled, so that the initial state persists for the complete web session. Hence, the static HMM is a non-dynamic model that still captures user heterogeneity. The static HMM is fed with both page and user covariates. A similar baseline was used for benchmarking in \citep{Ascarza.2018}.
\item MLSL is the HMM from \citet{Montgomery.2004}. The MLSL model assumes that the transition probabilities between the latent states are constant over time and across users. User heterogeneity in the emission is incorporated via covariates in the emission probabilities similar to our model. This model has been used earlier for clickstream modeling \citep{Ding.2015,Montgomery.2004}, and its model specification is widespread in marketing literature \citep[\eg,][]{Ascarza.2018,Netzer.2008}. A comparison of the theoretical properties between the MLSL and our model is provided in \Cref{sec:comparison_MLSL}. The MLSL model is specified in exactly same the way as our \model, yet it lacks a duration dependence in the latent state dynamics. Hence, by comparing it to our \model, we can identify the relative performance gain from including duration dependence. Reassuringly, we emphasize that the MLSL model has access to the exact same data as our \model, including the same covariates in the emission. 
\item A variant of the MLSL model that incorporates unobserved heterogeneity into the transition probabilities (denoted as MLSL + UH). We do so to check whether the differences in transition probabilities can be explained by individual-level unobserved heterogeneity. 
\end{enumerate}

\section{Results}
\label{sec:results}

\subsection{Prediction Performance in Detecting Exits}

We compare the different models in terms of predicting user exits without a purchase. The results are shown in \Cref{tbl:model_selection}. 
To start with, we first have to decide upon an appropriate number of latent states $K$. For this purpose, we estimate different models with varying $K$ based on data from the training set and then compare the model fit. Based on this procedure, we find that our \model performs best with three states. Hence, this model is also used in all subsequent sections where we report detailed estimation results for all parameters. We proceed analogously for all benchmark HMMs. Based on in-sample AUROC, the MLSL model (which is the HMM in \citet{Montgomery.2004}) is favored when choosing two states. This finding is consistent with prior literature \citep{Ding.2015,Montgomery.2004}.  

Next, we compare the prediction performance for unseen data (out-of-sample) based on the test set. Across all metrics, the \model yields the best performance. It achieves an AUROC of 0.7176, an AUPRC of 0.1364, and a hit rate of 0.5844. In particular, it outperforms the MLSL model by a considerable margin: our \model improves the AUROC by 7.65 percentage points, the AUPRC by 1.43 percentage points, and the hit rate by a full 18.18 percentage points. Hence, out of 100 user exits without a purchase, our model correctly identifies an additional 18 of them. Recall that both models have access to the same data; therefore, the performance gain must be solely attributed to the duration dependence. We also report confidence intervals (credible intervals). These are non-overlapping and, hence, imply that the performance improvement is not due to chance but is robust.

Accounting for unobserved heterogeneity in the transition probabilities in the MLSL (MLSL + UH) does not improve the prediction performance over the standard MLSL. Evidently, the unobserved heterogeneity is not able to capture the duration dependence in the transition probabilities. We also implemented a variant of our \model which accounts for unobserved heterogeneity. Although this variant performs well on the training data, it does perform poor on the test set due to overfitting. We present the corresponding results and model details in Appendix~\ref{appx:sec:DDHMM_UH}.

\begin{table}
\SingleSpacedXI
\footnotesize
\centering
\captionof{table}{Prediction Performance in Detecting User Exits.\label{tbl:model_selection}}
\begin{tabular}{lc ccc @{\hspace{1cm}} ccc}  
\toprule
&& \multicolumn{3}{c}{In-sample (training set)} &\multicolumn{3}{c}{Out-of-sample (test set)} \\
\cmidrule(r){3-5}
\cmidrule(r){6-8}
 Model & \#States & {AUROC} & {AUPRC} & {Hit rate} & {AUROC} & {AUPRC} & {Hit rate}\\
\midrule
Logistic regression  & {---} & 0.5995 & 0.0958 & 0.3812 & 0.5835 & 0.0855 & 0.4988\\
 &  & \scriptsize [0.59; 0.61]& \scriptsize[0.09; 0.10] & \scriptsize[0.37; 0.39] & \scriptsize[0.57; 0.59] & \scriptsize[0.08; 0.09] & \scriptsize[0.49; 0.50]\\[0.4em]
Random forest  & {---} & 0.6687 & 0.1182 & 0.4581 & 0.6585 &  0.1074 & 0.4988 \\
 &  & \scriptsize [0.66; 0.67]& \scriptsize[0.11; 0.12] & \scriptsize[0.45; 0.47] & \scriptsize[0.65; 0.66] & \scriptsize[0.10; 0.12] & \scriptsize[0.49; 0.50]\\[0.4em]
Markov model  & {---} & 0.5928 & 0.1644 & 0.3001 & 0.5848&  0.1536 & 0.2659 \\
 &  & \scriptsize [0.58; 0.60]& \scriptsize[0.16; 0.17] & \scriptsize[0.29 ; 0.31 ] & \scriptsize[0.57; 0.59] & \scriptsize[0.15; 0.16] & \scriptsize[0.26; 0.27]\\[0.4em]
LSTM  & {---}  & 0.6856 & 0.1335  & 0.4643 & 0.6787 & 0.1287 & 0.5766\\
 &  & \scriptsize [0.66; 0.70]& \scriptsize[0.12; 0.15] & \scriptsize[0.44; 0.48] & \scriptsize[0.65; 0.69] & \scriptsize[0.11; 0.14] & \scriptsize[0.55; 0.59]\\[0.4em]
M3PP  & & 0.6893 & 0.1245 & 0.4834 & 0.6912 & 0.1304 & 0.5793\\
 &  & \scriptsize [0.67; 0.70]& \scriptsize[0.09; 0.13] & \scriptsize[0.46; 0.49] & \scriptsize[0.67; 0.70] & \scriptsize[0.11; 0.14] & \scriptsize[0.55; 0.59]\\[0.4em]
Static HMM & 2 & 0.5298 & 0.0685  & 0.3059 & 0.5279 & 0.0649 & 0.3117\\ 
 &  & \scriptsize [0.52; 0.54]& \scriptsize[0.06; 0.07] & \scriptsize[0.29; 0.32] & \scriptsize[0.52; 0.53] & \scriptsize[0.06; 0.07] & \scriptsize[0.30; 0.32]\\[0.4em]
&3 & 0.5634 & 0.0816 & 0.3408 &0.5439 & 0.0724 & 0.3544\\ 
 &  & \scriptsize [0.55; 0.57]& \scriptsize[0.07; 0.09] & \scriptsize[0.06; 0.07] & \scriptsize[0.54; 0.55] & \scriptsize[0.03; 0.59] & \scriptsize[0.34; 0.36]\\[0.4em]
MLSL & 2 & 0.6597 & 0.1282  & 0.4795 & 0.6411 & 0.1221 & 0.4026\\ 
 &  & \scriptsize [0.64; 0.67]& \scriptsize[0.12; 0.13] & \scriptsize[0.47;0.49] & \scriptsize[0.63; 0.65] & \scriptsize[0.11; 0.13] & \scriptsize[0.39; 0.41]\\[0.4em]
 &3 & 0.6345 & 0.1186 & 0.4018 &0.6394 & 0.1242 & 0.4545\\
  &  & \scriptsize [0.63; 0.65]& \scriptsize[0.11; 0.13] & \scriptsize[0.39; 0.42] & \scriptsize[0.63; 0.64] & \scriptsize[0.12; 0.14] & \scriptsize[0.43; 0.46]\\[0.4em]
MLSL + UH & 2 & 0.5965 & 0.1224  & 0.4384 & 0.6057 & 0.1276 & 0.4026\\ 
 &  & \scriptsize [0.59; 0.61]& \scriptsize[0.12; 0.13] & \scriptsize[0.43; 0.45] & \scriptsize[0.59; 0.62] & \scriptsize[0.12; 0.14] & \scriptsize[0.38; 0.41]\\[0.4em]
&3 & 0.5978 & 0.1026 & 0.4416 & 0.6110 & 0.1506 & 0.4415\\
 &  & \scriptsize [0.58; 0.61]& \scriptsize[0.09; 0.11] & \scriptsize[0.43; 0.45] & \scriptsize[0.60; 0.62] & \scriptsize[0.14; 0.16] & \scriptsize[0.43; 0.45]\\[0.4em]
\midrule
Proposed \model &2 & 0.6987 & 0.1225 & 0.4795&0.6312 & 0.0897 & 0.3247\\ 
 &  & \scriptsize [0.68; 0.70]& \scriptsize[0.12; 0.13] & \scriptsize[0.47; 0.49] & \scriptsize[0.62; 0.64] & \scriptsize[0.08; 0.10] & \scriptsize[0.31; 0.34]\\[0.4em]
  & 3 & \bfseries 0.7033& \bfseries 0.1191 & \bfseries 0.4886 & \bfseries 0.7176 & \bfseries 0.1364 &  \bfseries 0.5844 \\ 
 &  & \scriptsize [0.69; 0.73]& \scriptsize[0.11; 0.13] & \scriptsize[0.48; 0.50] & \scriptsize[0.71; 0.72] & \scriptsize[0.13; 0.14] & \scriptsize[0.57; 0.59]\\[0.4em]
  &4 & 0.6152 & 0.1074& 0.3973 &0.6112 & 0.1214 & 0.3766\\ 
   &  & \scriptsize [0.60; 0.62]& \scriptsize[0.09; 0.11] & \scriptsize[0.37; 0.41] & \scriptsize[0.60; 0.619] & \scriptsize[0.12; 0.13] & \scriptsize[0.35; 0.38]\\
\bottomrule
\multicolumn{8}{p{15.5cm}}{\emph{Note:} Model selection is done based on the best AUROC on the training set (in-sample). MLSL is the HMM in \citet{Montgomery.2004}. MLSL + UH is the same as MLSL but, in addition, accounts for unobserved heterogeneity in the transition mechanism. For HMMs, credible intervals are used, otherwise confidence intervals (over 10 runs). Note that the confidence and credible intervals are non-overlapping. Higher values are better. Best model is highlighted in bold.}
\end{tabular}
\end{table}

\subsection{Estimation Results}
\label{sec:empiricalresults}

This section reports detailed estimation results for all components of the \model. These are provided for the variant using $K = 3$ states as this choice is favored during the above model selection.

\subsubsection{Characterization of Latent States.}
\label{sec:charlatentstates}

The \model returns three latent states with different characteristics in terms of both user intent and engagement. Consistent with \citep{Ding.2015,Montgomery.2004}, we thus label the states as: (i)~{goal-directed}, (ii)~``sticky'' browsing, and (iii)~``non-sticky'' browsing. The rationale for this naming is as follows:
\begin{itemize}
\item {{Goal-directed}:} This state is characterized by a high probability of purchase. When in this state, users spend little time on \textsc{Overview} pages but frequently proceed to checkout. As one would expect with directed behavior, the state is of short duration. If this state is left early, users are likely to continue with ``sticky'' browsing, during which they would collect further information. However, if the {goal-directed} state is left after a longer duration, users are likely to continue with ``non-sticky'' browsing, where they have an elevated risk of exiting.  
\item {``Sticky'' browsing:} In this state, user behavior shows a tendency towards extensive information collection. This is reflected by a high probability of visiting \textsc{Overview} and \textsc{Product} pages. In contrast to this, the risk of exit is close to zero. At the same time, this state reflects a high engagement level and is attained for a long duration. Hence, users in a ``sticky'' browsing state have a strong propensity towards renewing this state, largely independent of the previous duration in this state. Upon leaving the state, users adapt their intent based on the collected information: the more time a user has spent browsing, the more likely she is to transition to the {goal-directed} state. 
\item {``Non-sticky'' browsing:} The ``non-sticky'' browsing state also features product pages to a large extent as in experiential search. However, in contrast to ``sticky'' browsing, it is characterized by a lower engagement level, \ie, a comparatively short duration. Put differently, users are unlikely to renew this state, and there is a strong tendency towards switching to a different state or exiting altogether.
\end{itemize}

\subsubsection{Emission Probabilities.}
\label{sec:statedepbehavior}

The emission probabilities describe the state-dependent behavior, that is, the probability of observing a page $O_{it}$ conditional on a latent state. \Cref{tbl:Emission_average} reports these estimates. The probabilities are obtained by taking the average over all covariates inside the logit model. A detailed overview of all estimated parameters inside the logit (\ie, coefficients $\gamma_{is}^o$ and $\beta_{is}^o$) is located in Appendix~\ref{appx:sec:params}.

In the {goal-directed} state, users are likely to complete a purchase. The probability of a user reaching a \textsc{Checkout} page amounts to 9\,\%, which is higher than for any other state. This substantiates the naming in the literature \citep{Ding.2015,Montgomery.2004} in that users have a clear goal in mind. Furthermore, users frequently visit landing pages such as \textsc{Home} or \textsc{Account} (\eg, for tracking current shipments). In contrast, both \textsc{Overview} and \textsc{Product} pages are largely absent. In the ``sticky'' browsing state, users predominantly visit \textsc{Overview} and \textsc{Product} pages that foster experiential search. On these pages, users can collect information on product offerings and product details, respectively. The probabilities of both observing purchases (\ie, \textsc{Checkout}) and exits are estimated to be almost zero in this state. In the ``non-sticky'' browsing state, users visit landing pages (\ie, \textsc{Home} and \textsc{Account}) or pages that summarize purchases after checkout has been completed (\textsc{Order}), whereas the probability of making a purchase (\textsc{Checkout}) is almost zero. The ``non-sticky'' browsing state further entails a heightened probability of exits without purchase, which amounts to $\mathbb{P}$(\textsc{Exit})=\SI{65.5}{\percent}.

\begin{table}
\SingleSpacedXI
\footnotesize
\centering
\caption{Emission Probabilities Describing the State-Dependent Behavior.\label{tbl:Emission_average}}
\begin{tabular}{lccccccccccc}  
\toprule
Latent state & \multicolumn{9}{c}{Page $O_{it} \in \mathcal{O}$} \\
\cmidrule(lr){2-10}
& {\textsc{Home}} & {\textsc{Account}} & {\textsc{Overview}} & {\textsc{Prod.}} & {\textsc{Mktg. Page}} & {\textsc{Comm.}} & {\textsc{Checkout}} & {\textsc{Order}} & {\textsc{Exit}}  \\
\midrule
\csname @@input\endcsname Emission_average 
\bottomrule
\multicolumn{10}{l}{Stated: posterior means of emission probabilities (with average covariates); probabilities above 0.20 in bold.}
\end{tabular}
\end{table}
{}

\subsubsection{Initial Latent State Distribution.}

The initial distribution over latent states, $\pi$, reflects the probability of a user beginning her web session with a certain behavior. The estimated probabilities are listed in \Cref{tbl:Initial probability transition}. Evidently, most users (40\,\%) start their web session in a ``sticky'' browsing state. Taking both browsing states together, we find that a large share of users starts their web sessions with experiential activities and thus with information collection. In comparison, only 24\,\% of all users start their web session in a goal-directed state. 

\begin{table}
\SingleSpacedXI
\footnotesize
\centering
\caption{Initial State Distribution.\label{tbl:Initial probability transition}}
\begin{tabular}{lcc}  
\toprule
Latent state  & Estimate  & 95\,\% CI \\
\midrule
\csname @@input\endcsname initial
\bottomrule
\multicolumn{3}{l}{Stated: posterior mean; largest probability in bold}
\end{tabular}
\end{table}

\subsubsection{Distribution of Latent State Durations.}

\Cref{tbl:sojourn_time_coef} reports the expected duration of latent states. Both states -- {goal-directed} search and ``non-sticky'' browsing -- possess a short expected state duration. Their durations amount to $\mathbb{E}[D_s] = 2.39$ and $\mathbb{E}[D_s] = 1.26$ time periods, respectively. In contrast, a long latent state duration is observed for the ``sticky'' browsing state, where the duration amounts to $\mathbb{E}[D_s] = 5.98$ time periods.

Besides the latent state duration, \Cref{tbl:sojourn_time_coef} also provides the coefficients from the Weibull distribution. These control the overall magnitude of how \textquote{sticky} states are ($\theta_s$) and the corresponding prolongation of the \textquote{stickiness} along the time dimension ($c_s$). The latter is of particular interest as it lends to a distorted sense of time. We observe that the goal-directed state possesses a $c_s$ smaller than $1$. A value $c_s<1$ yields an increasing renewal probability. Therefore, the longer users remain in a state with $c_s<1$, the larger the renewal probability becomes. Hence, a user in a {goal-directed} state has a tendency to continue engaging in goal-directed behavior. A different observation is made for browsing. The ``sticky'' browsing has a $c_s$ close to 1. Hence, its renewal probability remains fairly constant over time. In contrast, the ``non-sticky'' browsing state possesses a $c_s$ larger than $1$. For $c_s > 1$, the resulting renewal probability is decreasing in the state duration $d$. As a result, the longer users remain in this state, the smaller the renewal probability becomes. In sum, the ``sticky'' browsing state is sticky over time as one would expect in a deep engagement, while the opposite is observed for ``non-sticky'' browsing. 

\begin{table}
\SingleSpacedXI
\footnotesize
\centering
\caption{Estimated Latent State Durations.\label{tbl:sojourn_time_coef}}
\begin{tabular}{lcccc}  
\toprule
Latent state & \multicolumn{2}{c}{Weibull distribution} & Latent state duration \\
\cmidrule(lr){2-3}\cmidrule(lr){4-4}
 &$\theta_s$ & $c_s$  &$\mathbb{E}[D_s]$ \\
\midrule
\csname @@input\endcsname duration_parameters 
\bottomrule
\multicolumn{3}{l}{Stated: posterior means with 95\,\% CI in parentheses}
\end{tabular}
\end{table}

\begin{figure}
\SingleSpacedXI
    \centering
	\caption{Renewal Probabilities as a Function of the Latent State Duration $d$ (\# pages).\label{fig:renewalprobability}}
    \input{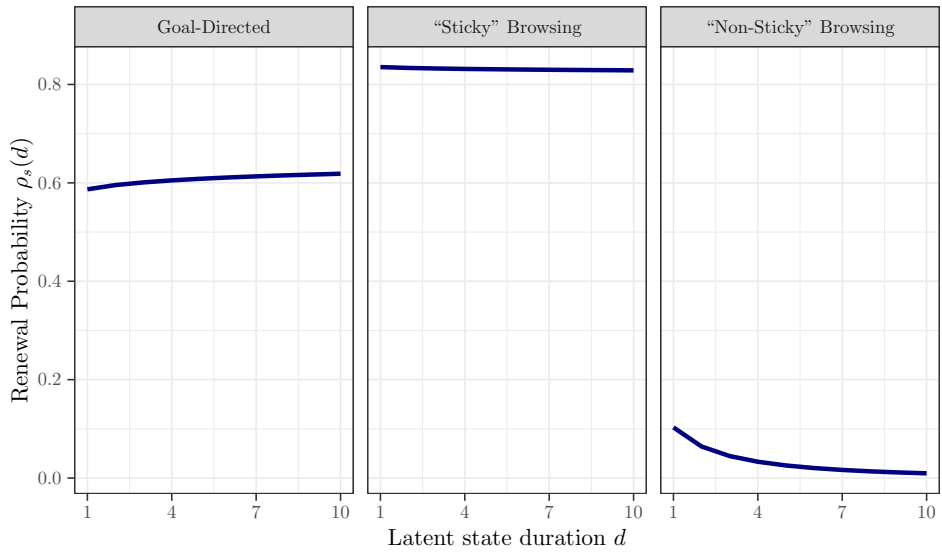}
\end{figure}

For better interpretability, we follow our earlier analysis and compute the renewal probability as a derived quantity in order to shed further light on how users arrive at a certain latent state duration. The renewal probability quantifies the relative likelihood of a user continuing in the current state for an additional time period (see \Cref{sec:theorymodel} for a mathematical derivation). The renewal probability is depicted in \Cref{fig:renewalprobability}. For the goal-directed state, the renewal probability is around 0.6 and remains fairly duration-independent. For ``sticky'' browsing, the renewal probability is substantially larger, which is attributed to the deep engagement experienced by the user and thus results in a longer latent state duration. A different pattern is obtained for ``non-sticky'' browsing. Here we observe an overall low renewal probability. Moreover, the renewal probability declines with $d$. After only two page visits in this state, the probability drops to a mere half of the initial value. In other words, users transition quickly to a different state.  

\subsubsection{Duration-Dependent Transition Mechanism.}

\begin{figure}
\caption{Duration-Dependent Transition Probabilities as a Function of the Latent State Duration $d$ (\ie, Number of Pages).\label{fig:tps}}
\begin{minipage}{.5\linewidth}
\centering
\scalebox{0.5}{
\begin{tikzpicture}[x=1pt,y=1pt]
\definecolor{fillColor}{RGB}{255,255,255}
\begin{scope}
\definecolor{drawColor}{RGB}{255,255,255}
\definecolor{fillColor}{RGB}{255,255,255}

\end{scope}
\begin{scope}
\definecolor{fillColor}{RGB}{255,255,255}

\definecolor{drawColor}{gray}{0.92}




\path[draw=drawColor,line width= 0.3pt,line join=round] ( 75.11, 27.01) --
	( 75.11,196.16);

\path[draw=drawColor,line width= 0.3pt,line join=round] (141.22, 27.01) --
	(141.22,196.16);

\path[draw=drawColor,line width= 0.3pt,line join=round] (207.33, 27.01) --
	(207.33,196.16);

\path[draw=drawColor,line width= 0.3pt,line join=round] (273.44, 27.01) --
	(273.44,196.16);

\path[draw=drawColor,line width= 0.6pt,line join=round] ( 60, 39.34) --
	(319.71, 39.34);

\path[draw=drawColor,line width= 0.6pt,line join=round] ( 60, 87.50) --
	(319.71, 87.50);

\path[draw=drawColor,line width= 0.6pt,line join=round] ( 60,135.66) --
	(319.71,135.66);

\path[draw=drawColor,line width= 0.6pt,line join=round] ( 60,183.83) --
	(319.71,183.83);


\path[draw=drawColor,line width= 0.6pt,line join=round] (108.16, 27.01) --
	(108.16,196.16);

\path[draw=drawColor,line width= 0.6pt,line join=round] (174.27, 27.01) --
	(174.27,196.16);

\path[draw=drawColor,line width= 0.6pt,line join=round] (240.38, 27.01) --
	(240.38,196.16);

\path[draw=drawColor,line width= 0.6pt,line join=round] (306.49, 27.01) --
	(306.49,196.16);
\definecolor{drawColor}{RGB}{0,0,128}

\path[draw=drawColor,line width= 1.7pt,line join=round] ( 
	( 68.50,113.92) --
	( 94.94,103.56) --
	(121.38, 93.32) --
	(147.83, 83.34) --
	(174.27, 73.76) --
	(200.72, 64.68) --
	(227.16, 56.19) --
	(253.60, 48.34) --
	(280.05, 41.18) --
	(306.49, 34.69);

\path[draw=drawColor,line width= 1.7pt,dash pattern=on 2pt off 2pt ,line join=round] ( 
	( 68.50,109.25) --
	( 94.94,119.60) --
	(121.38,129.84) --
	(147.83,139.82) --
	(174.27,149.40) --
	(200.72,158.48) --
	(227.16,166.97) --
	(253.60,174.82) --
	(280.05,181.99) --
	(306.49,188.47);
\definecolor{drawColor}{gray}{0.20}

\path[draw=drawColor,line width= 0.6pt,line join=round,line cap=round] ( 60, 27.01) rectangle (319.71,196.16);
\end{scope}
\begin{scope}
\path[clip] ( 60,196.16) rectangle (319.71,211.31);
\definecolor{drawColor}{gray}{0.20}
\definecolor{fillColor}{gray}{0.85}

\path[draw=drawColor,line width= 0.6pt,line join=round,line cap=round,fill=fillColor] ( 60,196.16) rectangle (319.71,211.31);
\definecolor{drawColor}{gray}{0.10}

\node[text=drawColor,anchor=base,inner sep=0pt, outer sep=0pt, scale=  0.64] at (185,201.53) {From Goal-Directed};
\end{scope}
\begin{scope}
\definecolor{drawColor}{gray}{0.20}

\path[draw=drawColor,line width= 0.6pt,line join=round] ( 68.50, 24.26) --
	( 68.50, 27.01);

\path[draw=drawColor,line width= 0.6pt,line join=round] (108.16, 24.26) --
	(108.16, 27.01);

\path[draw=drawColor,line width= 0.6pt,line join=round] (174.27, 24.26) --
	(174.27, 27.01);

\path[draw=drawColor,line width= 0.6pt,line join=round] (240.38, 24.26) --
	(240.38, 27.01);

\path[draw=drawColor,line width= 0.6pt,line join=round] (306.49, 24.26) --
	(306.49, 27.01);
\end{scope}
\begin{scope}
\path[clip] (  0.00,  0.00) rectangle (325.21,216.81);
\definecolor{drawColor}{gray}{0.30}

\node[text=drawColor,anchor=base,inner sep=0pt, outer sep=0pt, scale=  0.64] at ( 68.50, 17.65) {1};

\node[text=drawColor,anchor=base,inner sep=0pt, outer sep=0pt, scale=  0.64] at (110.49, 17.65) {2.5};

\node[text=drawColor,anchor=base,inner sep=0pt, outer sep=0pt, scale=  0.64] at (175.87, 17.65) {5};

\node[text=drawColor,anchor=base,inner sep=0pt, outer sep=0pt, scale=  0.64] at (241.26, 17.65) {7.5};

\node[text=drawColor,anchor=base,inner sep=0pt, outer sep=0pt, scale=  0.64] at (306.64, 17.65) {10};
\end{scope}
\begin{scope}
\definecolor{drawColor}{gray}{0.30}

\node[text=drawColor,anchor=base east,inner sep=0pt, outer sep=0pt, scale=  0.64] at ( 55, 37.13) {0.2};

\node[text=drawColor,anchor=base east,inner sep=0pt, outer sep=0pt, scale=  0.64] at ( 55, 85.30) {0.4};

\node[text=drawColor,anchor=base east,inner sep=0pt, outer sep=0pt, scale=  0.64] at ( 55,133.46) {0.6};

\node[text=drawColor,anchor=base east,inner sep=0pt, outer sep=0pt, scale=  0.64] at ( 55,181.62) {0.8};
\end{scope}
\begin{scope}
\path[clip] (  0.00,  0.00) rectangle (325.21,216.81);
\definecolor{drawColor}{gray}{0.20}

\path[draw=drawColor,line width= 0.6pt,line join=round] ( 59, 39.34) --
	( 61, 39.34);

\path[draw=drawColor,line width= 0.6pt,line join=round] ( 59, 87.50) --
	( 61, 87.50);

\path[draw=drawColor,line width= 0.6pt,line join=round] ( 59,135.66) --
	( 61,135.66);

\path[draw=drawColor,line width= 0.6pt,line join=round] ( 59,183.83) --
	( 61,183.83);
\end{scope}
\begin{scope}
\definecolor{drawColor}{RGB}{0,0,0}

\node[text=drawColor,anchor=base,inner sep=0pt, outer sep=0pt, scale=  0.80] at (174.27,  7.44) {Latent state duration $d$};
\end{scope}
\begin{scope}
\definecolor{drawColor}{RGB}{0,0,0}

\node[text=drawColor,rotate= 90.00,anchor=base,inner sep=0pt, outer sep=0pt, scale=  0.80] at ( 30,111.58) {Transition Probability};
\end{scope}
\begin{scope}
\definecolor{drawColor}{RGB}{0,0,0}
\definecolor{fillColor}{RGB}{255,255,255}

\path[draw=drawColor,line width= 1.1pt,line join=round,line cap=round,fill=fillColor] (229.87,102.65) rectangle (317.30,137.42);
\end{scope}
\begin{scope}
\path[clip] (  0.00,  0.00) rectangle (325.21,216.81);
\definecolor{drawColor}{RGB}{0,0,0}

\node[text=drawColor,anchor=base west,inner sep=0pt, outer sep=0pt, scale=  0.60] at (235.37,128.20) {Transition to};
\end{scope}
\begin{scope}
\path[clip] (  0.00,  0.00) rectangle (325.21,216.81);
\definecolor{fillColor}{RGB}{255,255,255}

\path[fill=fillColor] (235.37,116.69) rectangle (256.71,125.23);
\end{scope}
\begin{scope}
\path[clip] (  0.00,  0.00) rectangle (325.21,216.81);
\definecolor{drawColor}{RGB}{0,0,128}

\path[draw=drawColor,line width= 1.7pt,line join=round] (237.50,120.96) -- (244.57,120.96);
\end{scope}
\begin{scope}
\path[clip] (  0.00,  0.00) rectangle (325.21,216.81);
\definecolor{fillColor}{RGB}{255,255,255}

\path[fill=fillColor] (245.37,108.15) rectangle (266.71,116.69);
\end{scope}
\begin{scope}
\path[clip] (  0.00,  0.00) rectangle (315.21,216.81);
\definecolor{drawColor}{RGB}{0,0,128}

\path[draw=drawColor,line width= 1.7pt,dash pattern=on 2pt off 2pt ,line join=round] (237.50,112.42) -- (244.57,112.42);
\end{scope}
\begin{scope}
\path[clip] (  0.00,  0.00) rectangle (315.21,216.81);
\definecolor{drawColor}{RGB}{0,0,0}

\node[text=drawColor,anchor=base west,inner sep=0pt, outer sep=0pt, scale=  0.60] at (248.00,119.58) {``Sticky'' browsing};
\end{scope}
\begin{scope}
\path[clip] (  0.00,  0.00) rectangle (315.21,216.81);
\definecolor{drawColor}{RGB}{0,0,0}

\node[text=drawColor,anchor=base west,inner sep=0pt, outer sep=0pt, scale=  0.60] at (248.00,111.04) {``Non-sticky'' browsing};
\end{scope}
\end{tikzpicture}}
\end{minipage}%
\begin{minipage}{.5\linewidth}
\centering
\scalebox{0.5}{
\begin{tikzpicture}[x=1pt,y=1pt]
\definecolor{fillColor}{RGB}{255,255,255}
\path[use as bounding box,fill=fillColor,fill opacity=0.00] (0,0) rectangle (325.21,216.81);
\begin{scope}
\definecolor{drawColor}{RGB}{255,255,255}
\definecolor{fillColor}{RGB}{255,255,255}

\end{scope}
\begin{scope}
\definecolor{fillColor}{RGB}{255,255,255}

\definecolor{drawColor}{gray}{0.92}




\path[draw=drawColor,line width= 0.3pt,line join=round] ( 75.11, 27.01) --
	( 75.11,196.16);

\path[draw=drawColor,line width= 0.3pt,line join=round] (141.22, 27.01) --
	(141.22,196.16);

\path[draw=drawColor,line width= 0.3pt,line join=round] (207.33, 27.01) --
	(207.33,196.16);

\path[draw=drawColor,line width= 0.3pt,line join=round] (273.44, 27.01) --
	(273.44,196.16);

\path[draw=drawColor,line width= 0.6pt,line join=round] ( 60, 39.34) --
	(319.71, 39.34);

\path[draw=drawColor,line width= 0.6pt,line join=round] ( 60, 87.50) --
	(319.71, 87.50);

\path[draw=drawColor,line width= 0.6pt,line join=round] ( 60,135.66) --
	(319.71,135.66);

\path[draw=drawColor,line width= 0.6pt,line join=round] ( 60,183.83) --
	(319.71,183.83);


\path[draw=drawColor,line width= 0.6pt,line join=round] (108.16, 27.01) --
	(108.16,196.16);

\path[draw=drawColor,line width= 0.6pt,line join=round] (174.27, 27.01) --
	(174.27,196.16);

\path[draw=drawColor,line width= 0.6pt,line join=round] (240.38, 27.01) --
	(240.38,196.16);

\path[draw=drawColor,line width= 0.6pt,line join=round] (306.49, 27.01) --
	(306.49,196.16);
\definecolor{drawColor}{RGB}{0,0,128}

\path[draw=drawColor,line width= 1.7pt,line join=round] ( 
	( 71.26,147.37) --
	( 97.41,162.39) --
	(123.57,172.73) --
	(149.72,179.33) --
	(175.87,183.32) --
	(202.03,185.66) --
	(228.18,187.01) --
	(254.33,187.79) --
	(280.49,188.22) --
	(306.64,188.47);

\path[draw=drawColor,line width= 1.7pt,dash pattern=on 2pt off 2pt ,line join=round] ( 
	( 71.26, 75.80) --
	( 97.41, 60.78) --
	(123.57, 50.43) --
	(149.72, 43.84) --
	(175.87, 39.85) --
	(202.03, 37.50) --
	(228.18, 36.15) --
	(254.33, 35.38) --
	(280.49, 34.94) --
	(306.64, 34.69);
\definecolor{drawColor}{gray}{0.20}

\path[draw=drawColor,line width= 0.6pt,line join=round,line cap=round] ( 60, 27.01) rectangle (319.71,196.16);
\end{scope}
\begin{scope}
\path[clip] ( 60,196.16) rectangle (319.71,211.31);
\definecolor{drawColor}{gray}{0.20}
\definecolor{fillColor}{gray}{0.85}

\path[draw=drawColor,line width= 0.6pt,line join=round,line cap=round,fill=fillColor] ( 60,196.16) rectangle (319.71,211.31);
\definecolor{drawColor}{gray}{0.10}

\node[text=drawColor,anchor=base,inner sep=0pt, outer sep=0pt, scale=  0.64] at (185,201.53) {From ``Sticky'' Browsing};
\end{scope}
\begin{scope}
\path[clip] (  0.00,  0.00) rectangle (325.21,216.81);
\definecolor{drawColor}{gray}{0.20}

\path[draw=drawColor,line width= 0.6pt,line join=round] ( 71.26, 24.26) --
	( 71.26, 27.01);

\path[draw=drawColor,line width= 0.6pt,line join=round] (110.49, 24.26) --
	(110.49, 27.01);

\path[draw=drawColor,line width= 0.6pt,line join=round] (175.87, 24.26) --
	(175.87, 27.01);

\path[draw=drawColor,line width= 0.6pt,line join=round] (241.26, 24.26) --
	(241.26, 27.01);

\path[draw=drawColor,line width= 0.6pt,line join=round] (306.64, 24.26) --
	(306.64, 27.01);
\end{scope}
\begin{scope}
\path[clip] (  0.00,  0.00) rectangle (325.21,216.81);
\definecolor{drawColor}{gray}{0.30}

\node[text=drawColor,anchor=base,inner sep=0pt, outer sep=0pt, scale=  0.64] at ( 68.50, 17.65) {1};

\node[text=drawColor,anchor=base,inner sep=0pt, outer sep=0pt, scale=  0.64] at (110.49, 17.65) {2.5};

\node[text=drawColor,anchor=base,inner sep=0pt, outer sep=0pt, scale=  0.64] at (175.87, 17.65) {5};

\node[text=drawColor,anchor=base,inner sep=0pt, outer sep=0pt, scale=  0.64] at (241.26, 17.65) {7.5};

\node[text=drawColor,anchor=base,inner sep=0pt, outer sep=0pt, scale=  0.64] at (306.64, 17.65) {10};
\end{scope}
\begin{scope}
\path[clip] (  0.00,  0.00) rectangle (325.21,216.81);
\definecolor{drawColor}{gray}{0.30}

\node[text=drawColor,anchor=base east,inner sep=0pt, outer sep=0pt, scale=  0.64] at ( 55, 37.13) {0.2};

\node[text=drawColor,anchor=base east,inner sep=0pt, outer sep=0pt, scale=  0.64] at ( 55, 85.30) {0.4};

\node[text=drawColor,anchor=base east,inner sep=0pt, outer sep=0pt, scale=  0.64] at ( 55,133.46) {0.6};

\node[text=drawColor,anchor=base east,inner sep=0pt, outer sep=0pt, scale=  0.64] at ( 55,181.62) {0.8};
\end{scope}
\begin{scope}
\path[clip] (  0.00,  0.00) rectangle (325.21,216.81);
\definecolor{drawColor}{gray}{0.20}

\path[draw=drawColor,line width= 0.6pt,line join=round] ( 59, 39.34) --
	( 61, 39.34);

\path[draw=drawColor,line width= 0.6pt,line join=round] ( 59, 87.50) --
	( 61, 87.50);

\path[draw=drawColor,line width= 0.6pt,line join=round] ( 59,135.66) --
	( 61,135.66);

\path[draw=drawColor,line width= 0.6pt,line join=round] ( 59,183.83) --
	( 61,183.83);
\end{scope}
\begin{scope}
\path[clip] (  0.00,  0.00) rectangle (325.21,216.81);
\definecolor{drawColor}{RGB}{0,0,0}

\node[text=drawColor,anchor=base,inner sep=0pt, outer sep=0pt, scale=  0.80] at (174.27,  7.44) {Latent state duration $d$};
\end{scope}
\begin{scope}
\path[clip] (  0.00,  0.00) rectangle (325.21,216.81);
\definecolor{drawColor}{RGB}{0,0,0}

\node[text=drawColor,rotate= 90.00,anchor=base,inner sep=0pt, outer sep=0pt, scale=  0.80] at ( 30,111.58) {Transition Probability};
\end{scope}
\begin{scope}
\path[clip] (  0.00,  0.00) rectangle (325.21,216.81);
\definecolor{drawColor}{RGB}{0,0,0}
\definecolor{fillColor}{RGB}{255,255,255}

\path[draw=drawColor,line width= 1.1pt,line join=round,line cap=round,fill=fillColor] (229.87,102.65) rectangle (317.30,137.42);
\end{scope}
\begin{scope}
\path[clip] (  0.00,  0.00) rectangle (325.21,216.81);
\definecolor{drawColor}{RGB}{0,0,0}

\node[text=drawColor,anchor=base west,inner sep=0pt, outer sep=0pt, scale=  0.60] at (235.37,128.20) {Transition to};
\end{scope}
\begin{scope}
\path[clip] (  0.00,  0.00) rectangle (325.21,216.81);
\definecolor{fillColor}{RGB}{255,255,255}

\path[fill=fillColor] (245.37,116.69) rectangle (266.71,125.23);
\end{scope}
\begin{scope}
\path[clip] (  0.00,  0.00) rectangle (325.21,216.81);
\definecolor{drawColor}{RGB}{0,0,128}

\path[draw=drawColor,line width= 1.7pt,line join=round] (237.50,120.96) -- (244.57,120.96);
\end{scope}
\begin{scope}
\path[clip] (  0.00,  0.00) rectangle (325.21,216.81);
\definecolor{fillColor}{RGB}{255,255,255}

\path[fill=fillColor] (245.37,108.15) rectangle (266.71,116.69);
\end{scope}
\begin{scope}
\path[clip] (  0.00,  0.00) rectangle (325.21,216.81);
\definecolor{drawColor}{RGB}{0,0,128}

\path[draw=drawColor,line width= 1.7pt,dash pattern=on 2pt off 2pt ,line join=round] (237.50,112.42) -- (244.57,112.42);
\end{scope}
\begin{scope}
\path[clip] (  0.00,  0.00) rectangle (325.21,216.81);
\definecolor{drawColor}{RGB}{0,0,0}

\node[text=drawColor,anchor=base west,inner sep=0pt, outer sep=0pt, scale=  0.60] at (248.0,119.58) {Goal-directed};
\end{scope}
\begin{scope}
\path[clip] (  0.00,  0.00) rectangle (325.21,216.81);
\definecolor{drawColor}{RGB}{0,0,0}

\node[text=drawColor,anchor=base west,inner sep=0pt, outer sep=0pt, scale=  0.60] at (248.0,111.04) {``Non-sticky'' browsing};
\end{scope}
\end{tikzpicture}}
\end{minipage}\par\medskip

\centering
\scalebox{0.5}{
\begin{tikzpicture}[x=1pt,y=1pt]
\definecolor{fillColor}{RGB}{255,255,255}
\path[use as bounding box,fill=fillColor,fill opacity=0.00] (0,0) rectangle (325.21,216.81);
\begin{scope}
\definecolor{drawColor}{RGB}{255,255,255}
\definecolor{fillColor}{RGB}{255,255,255}

\end{scope}
\begin{scope}
\definecolor{fillColor}{RGB}{255,255,255}

\definecolor{drawColor}{gray}{0.92}




\path[draw=drawColor,line width= 0.3pt,line join=round] ( 75.11, 27.01) --
	( 75.11,196.16);

\path[draw=drawColor,line width= 0.3pt,line join=round] (141.22, 27.01) --
	(141.22,196.16);

\path[draw=drawColor,line width= 0.3pt,line join=round] (207.33, 27.01) --
	(207.33,196.16);

\path[draw=drawColor,line width= 0.3pt,line join=round] (273.44, 27.01) --
	(273.44,196.16);

\path[draw=drawColor,line width= 0.6pt,line join=round] ( 60, 39.34) --
	(319.71, 39.34);

\path[draw=drawColor,line width= 0.6pt,line join=round] ( 60, 87.50) --
	(319.71, 87.50);

\path[draw=drawColor,line width= 0.6pt,line join=round] ( 60,135.66) --
	(319.71,135.66);

\path[draw=drawColor,line width= 0.6pt,line join=round] ( 60,183.83) --
	(319.71,183.83);


\path[draw=drawColor,line width= 0.6pt,line join=round] (108.16, 27.01) --
	(108.16,196.16);

\path[draw=drawColor,line width= 0.6pt,line join=round] (174.27, 27.01) --
	(174.27,196.16);

\path[draw=drawColor,line width= 0.6pt,line join=round] (240.38, 27.01) --
	(240.38,196.16);

\path[draw=drawColor,line width= 0.6pt,line join=round] (306.49, 27.01) --
	(306.49,196.16);
\definecolor{drawColor}{RGB}{0,0,128}

\path[draw=drawColor,line width= 1.7pt,line join=round] ( 
	( 71.26,122.50) --
	( 97.41,132.82) --
	(123.57,142.63) --
	(149.72,151.75) --
	(175.87,160.06) --
	(202.03,167.48) --
	(228.18,173.99) --
	(254.33,179.62) --
	(280.49,184.42) --
	(306.64,188.47);

\path[draw=drawColor,line width= 1.7pt,dash pattern=on 2pt off 2pt ,line join=round] ( 
	( 71.26,100.66) --
	( 97.41, 90.35) --
	(123.57, 80.54) --
	(149.72, 71.41) --
	(175.87, 63.10) --
	(202.03, 55.68) --
	(228.18, 49.17) --
	(254.33, 43.54) --
	(280.49, 38.74) --
	(306.64, 34.69);
\definecolor{drawColor}{gray}{0.20}

\path[draw=drawColor,line width= 0.6pt,line join=round,line cap=round] ( 60, 27.01) rectangle (319.71,196.16);
\end{scope}
\begin{scope}
\path[clip] ( 60,196.16) rectangle (319.71,211.31);
\definecolor{drawColor}{gray}{0.20}
\definecolor{fillColor}{gray}{0.85}

\path[draw=drawColor,line width= 0.6pt,line join=round,line cap=round,fill=fillColor] ( 60,196.16) rectangle (319.71,211.31);
\definecolor{drawColor}{gray}{0.10}

\node[text=drawColor,anchor=base,inner sep=0pt, outer sep=0pt, scale=  0.64] at (185,201.53) {From ``Non-Sticky'' Browsing};
\end{scope}
\begin{scope}
\path[clip] (  0.00,  0.00) rectangle (325.21,216.81);
\definecolor{drawColor}{gray}{0.20}

\path[draw=drawColor,line width= 0.6pt,line join=round] ( 71.26, 24.26) --
	( 71.26, 27.01);

\path[draw=drawColor,line width= 0.6pt,line join=round] (110.49, 24.26) --
	(110.49, 27.01);

\path[draw=drawColor,line width= 0.6pt,line join=round] (175.87, 24.26) --
	(175.87, 27.01);

\path[draw=drawColor,line width= 0.6pt,line join=round] (241.26, 24.26) --
	(241.26, 27.01);

\path[draw=drawColor,line width= 0.6pt,line join=round] (306.64, 24.26) --
	(306.64, 27.01);
\end{scope}
\begin{scope}
\path[clip] (  0.00,  0.00) rectangle (325.21,216.81);
\definecolor{drawColor}{gray}{0.30}

\node[text=drawColor,anchor=base,inner sep=0pt, outer sep=0pt, scale=  0.64] at ( 68.50, 17.65) {1};

\node[text=drawColor,anchor=base,inner sep=0pt, outer sep=0pt, scale=  0.64] at (110.49, 17.65) {2.5};

\node[text=drawColor,anchor=base,inner sep=0pt, outer sep=0pt, scale=  0.64] at (175.87, 17.65) {5};

\node[text=drawColor,anchor=base,inner sep=0pt, outer sep=0pt, scale=  0.64] at (241.26, 17.65) {7.5};

\node[text=drawColor,anchor=base,inner sep=0pt, outer sep=0pt, scale=  0.64] at (306.64, 17.65) {10};
\end{scope}
\begin{scope}
\path[clip] (  0.00,  0.00) rectangle (325.21,216.81);
\definecolor{drawColor}{gray}{0.30}

\node[text=drawColor,anchor=base east,inner sep=0pt, outer sep=0pt, scale=  0.64] at ( 55, 37.13) {0.2};

\node[text=drawColor,anchor=base east,inner sep=0pt, outer sep=0pt, scale=  0.64] at ( 55, 85.30) {0.4};

\node[text=drawColor,anchor=base east,inner sep=0pt, outer sep=0pt, scale=  0.64] at ( 55,133.46) {0.6};

\node[text=drawColor,anchor=base east,inner sep=0pt, outer sep=0pt, scale=  0.64] at ( 55,181.62) {0.8};
\end{scope}
\begin{scope}
\path[clip] (  0.00,  0.00) rectangle (325.21,216.81);
\definecolor{drawColor}{gray}{0.20}

\path[draw=drawColor,line width= 0.6pt,line join=round] ( 59, 39.34) --
	( 61, 39.34);

\path[draw=drawColor,line width= 0.6pt,line join=round] ( 59, 87.50) --
	( 61, 87.50);

\path[draw=drawColor,line width= 0.6pt,line join=round] ( 59,135.66) --
	( 61,135.66);

\path[draw=drawColor,line width= 0.6pt,line join=round] ( 59,183.83) --
	( 61,183.83);
\end{scope}
\begin{scope}
\path[clip] (  0.00,  0.00) rectangle (325.21,216.81);
\definecolor{drawColor}{RGB}{0,0,0}

\node[text=drawColor,anchor=base,inner sep=0pt, outer sep=0pt, scale=  0.80] at (174.27,  7.44) {Latent state duration $d$};
\end{scope}
\begin{scope}
\path[clip] (  0.00,  0.00) rectangle (325.21,216.81);
\definecolor{drawColor}{RGB}{0,0,0}

\node[text=drawColor,rotate= 90.00,anchor=base,inner sep=0pt, outer sep=0pt, scale=  0.80] at ( 30,111.58) {Transition Probability};
\end{scope}
\begin{scope}
\path[clip] (  0.00,  0.00) rectangle (325.21,216.81);
\definecolor{drawColor}{RGB}{0,0,0}
\definecolor{fillColor}{RGB}{255,255,255}

\path[draw=drawColor,line width= 1.1pt,line join=round,line cap=round,fill=fillColor] (229.87,102.65) rectangle (317.30,137.42);
\end{scope}
\begin{scope}
\path[clip] (  0.00,  0.00) rectangle (325.21,216.81);
\definecolor{drawColor}{RGB}{0,0,0}

\node[text=drawColor,anchor=base west,inner sep=0pt, outer sep=0pt, scale=  0.60] at (235.37,128.20) {Transition to};
\end{scope}
\begin{scope}
\path[clip] (  0.00,  0.00) rectangle (325.21,216.81);
\definecolor{fillColor}{RGB}{255,255,255}

\path[fill=fillColor] (245.37,116.69) rectangle (266.71,125.23);
\end{scope}
\begin{scope}
\path[clip] (  0.00,  0.00) rectangle (325.21,216.81);
\definecolor{drawColor}{RGB}{0,0,128}

\path[draw=drawColor,line width= 1.7pt,line join=round] (237.50,120.96) -- (244.57,120.96);
\end{scope}
\begin{scope}
\path[clip] (  0.00,  0.00) rectangle (325.21,216.81);
\definecolor{fillColor}{RGB}{255,255,255}

\path[fill=fillColor] (245.37,108.15) rectangle (266.71,116.69);
\end{scope}
\begin{scope}
\path[clip] (  0.00,  0.00) rectangle (325.21,216.81);
\definecolor{drawColor}{RGB}{0,0,128}

\path[draw=drawColor,line width= 1.7pt,dash pattern=on 2pt off 2pt ,line join=round] (237.50,112.42) -- (244.57,112.42);
\end{scope}
\begin{scope}
\path[clip] (  0.00,  0.00) rectangle (325.21,216.81);
\definecolor{drawColor}{RGB}{0,0,0}

\node[text=drawColor,anchor=base west,inner sep=0pt, outer sep=0pt, scale=  0.60] at (248.0,119.58) {Goal-directed};
\end{scope}
\begin{scope}
\path[clip] (  0.00,  0.00) rectangle (325.21,216.81);
\definecolor{drawColor}{RGB}{0,0,0}

\node[text=drawColor,anchor=base west,inner sep=0pt, outer sep=0pt, scale=  0.60] at (248.0,111.04) {``Sticky'' Browsing};
\end{scope}
\end{tikzpicture}}
\end{figure}
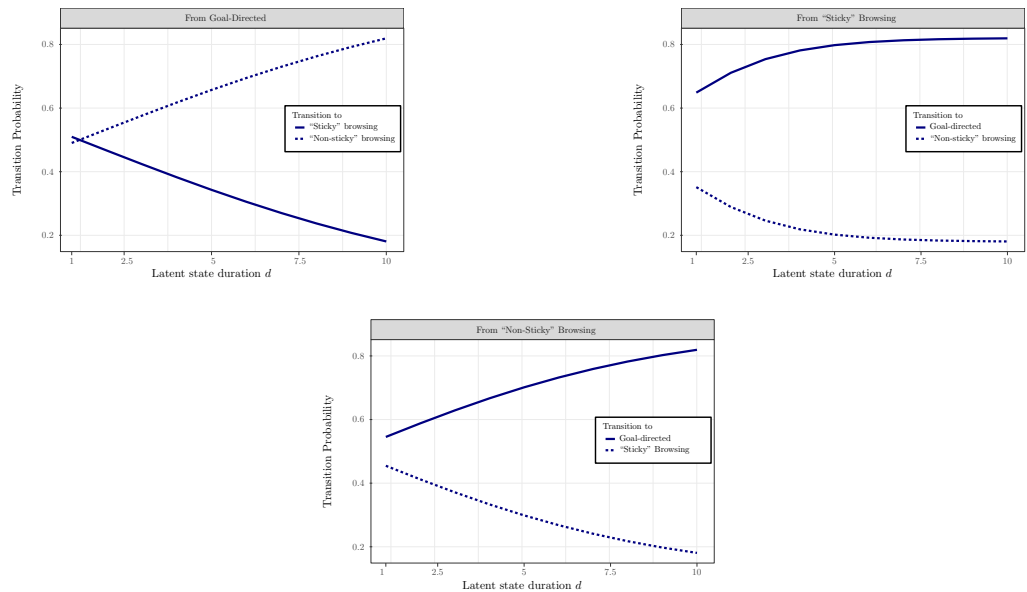

The transition mechanism describes the dynamics among states over time. Recall that self-transitions (\ie, transitions whereby the same state is continued) are already captured by the latent state durations. Thus, in the following, we focus only on transitions whereby one states switches into a different one. The transition probability is subject to duration dependence, as shown in \Cref{fig:tps}. As a result, transitions into some states can become more likely over time, whereas the propensity for others declines.   

For the {goal-directed} state, we yield mixed findings depending on the latent state duration. If users leave this state early, they are likely to continue with ``sticky'' browsing (\ie, to collect more information). If users leave their goal-directed state after a long duration, they are more likely to proceed with ``non-sticky'' browsing instead. The latter might be especially pronounced in cases in which users could not achieve their goal after a long duration, as a transition to ``non-sticky'' browsing is also associated with a substantially larger risk of exiting altogether. In the case of ``sticky'' browsing, users are more likely to transition to a {goal-directed} state than to ``non-sticky'' browsing. This pattern is reinforced by a longer duration: the more time users have spent performing information collection, the more probable it is that they have formed a specific goal and will pursue a {goal-directed} search. In ``non-sticky'' browsing, the effect is similar but less pronounced.

The dynamics in renewal probabilities and the dynamics in transition probabilities are two distinct concepts. The renewal probabilities are concerned with the probability of remaining in the current state. On the contrary, the transition probabilities are concerned with the probability of transitioning to another state once the user has decided to leave the current state. As such, these two concepts are concerned with two distinct parts of the user behavior. In particular, the relatively stable renewal probabilities in \Cref{fig:renewalprobability} indicate that the duration of a state does barely influence the decision to remain in the current state. Different to that, once the user decided to leave the current state, the duration of this state has a large influence on the next state the user transitions to (as in \Cref{fig:tps}). 

\subsection{Robustness Checks}
\label{sec::robustness_results}

We test the robustness of our results by performing a series of additional experiments. Specifically, we implemented a variety of model extensions using further time-specific covariates. Here we used: (i)~time of day (continuous); (ii)~time of day (categorical: morning, noon, evening); and (iii)~weekday (categorical: 0--5, where 0 denotes weekend). However, we found that including any of these covariates does not substantially change the estimation result but, if so, increased the risk of overfitting. Furthermore, we noted that including more covariates increased the estimation variance (i.e., wider confidence intervals). In addition, we tested the variant where covariates are included in the transition probabilities (instead of emission probabilities). However, this led to an inferior performance of the model. 

\section{Case Study with Dynamic Targeting}
\label{sec:case_study}

In this section, we demonstrate the value of our model to marketers. Recall that our partner company is positioned in a high-end customer segment, because of which traditional marketing interventions (\eg, coupons, price promotions) are currently not in use. Instead, our partner company recently introduced marketing pages with curated content that should guide customers toward purchasing. The marketing pages provide information on trending products and are often enriched with overviews on specific product categories. Due to positive feedback from customers, the partner company is currently investigating whether they should make even wider use of such curated pages. To this end, one idea is an update to a webpage where curated content is displayed dynamically during the web session when a user is at risk of exiting.

Informed by the above setting, we treat marketing pages as interventions for targeting customers. We then proceeded as follows. The marketing pages enter our model as observations because they represent pages in the web session (and not as covariates that enter the transition component). We use the estimated model parameter from \Cref{sec:empiricalresults} and then sample user sessions from the model parameters. Whenever our model predicts that a user is about to exit without a purchase, we simulate the web session that would occur if  a marketing page with curated content was dynamically displayed. The underlying rationale is that this marketing page should make the user's latent state transition to a goal-directed state, which is characterized by a heightened probability to purchase (see \Cref{tbl:Emission_average}). Furthermore, we investigate different effectiveness levels of this intervention (\ie, different probabilities of transitioning a user to a goal-directed state). Finally, we vary the underlying targeting strategy across two scenarios.

\emph{Scenario 1:} First, we consider the case when the effect of a marketing page on the latent state is uniform across different latent states. As such, the marketing page is identical in all latent state (without tailoring) and thus has the same effect on all latent states. 
\emph{Scenario 2:} It is likely that the effect of the intervention might depend on the latent state itself. Hence, we further consider the case in which the effect on the latent user state may vary across different states. As such, the marketing page can be tailored to the latent state, and, hence, if the latent state is known (as in latent state models), the effectiveness may be higher. In particular, such behavior arises naturally from our \model model since the HMM-based framework is widely used in marketing to segment the user base according to their latent state. 

\begin{table}
\SingleSpacedXI
\footnotesize
\centering
\captionof{table}{Case Study: Improvement of Conversion Rate from Marketing Interventions.}\label{tbl:case_study}
\begin{tabular}{lcc ccc ccc}  
\toprule
&& &\multicolumn{3}{c}{Scenario 1} &\multicolumn{3}{c}{Scenario 2} \\
\cmidrule(r){4-6}
\cmidrule(r){7-9}
 Model & \makecell[c]{Latent\\States} & \makecell[c]{Duration\\Dep.} &{(a)} & {(b)}  & {(c)} & {(a)} & {(b)} & {(c)}\\
\midrule
Logistic regression  & {\xmark} &{\xmark} & -0.1\,\%  & 7.5\,\%  & 16.0\,\% &  0.8\,\% & 6.5\,\%  & 13.3\,\%\\[0.4em]
Random forest  & {\xmark}&{\xmark}& 1.5\,\% & 4.0\,\% &  17.3\,\%&  2.6\,\%  & 7.8\,\% & 24.0\,\% \\[0.4em]
Markov model  & {\xmark}&{\xmark} & -2.4\,\%  &-3.1\,\% &  7.2\,\% &  0.0\,\%  & 1.0\,\%  & -1.2\,\%\\[0.4em]
LSTM  & {\xmark}&{\xmark}& 0.2\,\% & 1.2\,\% & 43.4\,\%&  -1.3\,\%  & 7.0\,\%  & 13.0\,\%\\[0.4em]
Static HMM & {\cmark}&{\xmark}  & 0.6\,\%  & 3.2\,\%  &  8.6\,\%&  1.7\,\%  & 5.7\,\%  & 1.2\,\%\\[0.4em]
MLSL & {\cmark}&{\xmark} &  -1.6\,\% & 4.3\,\%  &  12.6\,\%&  1.1\,\%  & 10.8\,\%  & 10.9\,\%\\[0.4em]
\midrule
Proposed \model &{\cmark}&{\cmark}& \bf{5.5\,\%}  & \bf{10.5\,\%}  &  \bf{47.0\,\%}  &  \bf{6.0\,\%}  &\bf{25.4\,\%}  &\bf{45.1\,\%}\\
\bottomrule
\multicolumn{9}{p{13.5cm}}{\emph{Note:} Model selection is done based on the best AUROC as before (see \Cref{tbl:model_selection}). For each scenario, we investigate 3 variants: (a)~10\,\% intervention effectiveness, (b)~50\,\% intervention effectiveness, and  (c)~100\,\% intervention effectiveness, where intervention effectiveness describes the probability that a user transitions to a goal-directed state, when intervened with a marketing page. Best model is highlighted in bold.}
\end{tabular}
\end{table}
 
In \Cref{tbl:case_study}, we present the results. In particular, for each scenario, the variants (a), (b), and (c) refer to different levels of effectiveness of the marketing page in transitioning the user to a goal-direct state.  For each scenario, we investigate 3 variants: (a)~10\,\% intervention effectiveness, (b)~50\,\% intervention effectiveness, and  (c)~100\,\% intervention effectiveness, where intervention effectiveness describes the probability that a user transitions to a goal-directed state, when intervened with a marketing page. In scenario~2, the effectiveness further depends on the latent state, and, hence, it can only be leverage when using latent state models. Here we further consider the effectiveness to be reduced by half if the marketing page cannot be tailored to the latent state (which requires a latent state model).

Based on \Cref{tbl:case_study}, we make the following two observations. First, our \model achieves a clear improvement in the conversion rate. Second, some benchmarks even decreases the conversion rate in some scenarios. This is due to the inferior prediction performance of these models. In particular, even if some exiting users are detected because of the false positive rate, we would also intervene on some users that would have purchased but did not purchase once we intervened. Hence, if the false positive rate is too high compared to the hit rate, the conversion rate even decreases. This indicates the potential benefit of our \model for detecting users at the risk of exiting without a purchase and, as such, improving the conversion rate of an e-commerce retailer through displaying marketing pages with curated content. In sum, dynamically targeting customers with curated content is particularly effective when our proposed HMM with duration dependence is used by marketers.

\section{Discussion}
\label{sec:discussion} 

\subsection{Implications for Marketers}

Our work is helpful to marketers from e-commerce websites. E-commerce websites are confronted with the problem that most users exit without a purchase \citep{Statista.2019}. This has a deleterious effect on prospect of sales, and, in order to steer users towards purchasing, marketers require models that identify users at risk of exiting. Our work follows the prediction task of \citet{Ding.2015} and \citet{Montgomery.2004}, but offers an approach whereby exits without a purchase can be detected more accurately. By using our model, marketers can make correct predictions for an additional 18 (out of 100) users at risk of exiting. Overall, our model allows marketers to detect no-purchase exits with a superior AUROC of 0.7176.

With the help of exit predictions, marketers can allocate resources that steer users towards making a purchase. This is best demonstrated based on the following example. Let us assume that an e-commerce website has a typical conversion rate of \SI{4}{\%} \citep{Statista.2019}. In other words, out of 100 users, 4 users make a purchase and 96 exit without a purchase. Recall that our \model can correctly detect 18 additional exits without a purchase as compared to existing HMMs for that task \citep{Montgomery.2004}. Hence, if marketing efforts convert even one of the 18 exits into a purchase, this would yield a total of 5 purchases (out of 100 users) and thus bolster the conversion rate by \SI{25}{\%}.

Marketers can use our model to segment their customer base not only by user intents but also by engagement levels and then adapt their marketing efforts accordingly. When in a state of ``sticky'' browsing, users are unlikely to leave the website as they enjoy their experience. Hence, there is comparatively little need for marketing efforts that prevent users from exiting. Instead, promising marketing efforts include those that facilitate information collection (\eg, chatbots) or drive users towards making a purchase (\eg, price promotions). Marketing efforts can also be varied by the duration of latent states. For instance, triggering stimuli after a long latent state duration will target customers who are already likely to continue with goal-directed behavior. When users are in a state of ``non-sticky'' browsing, marketing efforts are needed that prevent them from exiting. Otherwise, there is a substantial risk of exit (\ie, the exit probability amounts to 65.5\,\% per visited page). On top of that, this state is attained only for a short duration, and, hence, early interventions are important. In our case study, we have shown above how our model can help marketers in generating managerial insights (see \Cref{sec:case_study}). Therein, we have analyzed the role of marketing pages with curated content in driving conversion. 

\subsection{Implications for Research}

Our work connects to earlier research on modeling online behavior \citep[\eg,][]{Bucklin.2003,moe2003buying,Moe.2004,Moe.2012,Sismeiro.2004}. Previous literature has developed mathematical models for distinguishing different user intents (\ie, goal-directed vs. browsing) in web sessions \citep{Ding.2015,Montgomery.2004}. In this paper, we also find differences between ``sticky'' and ``non-sticky'' states, which allows us to account for a distorted sense of time and which thus shares similarities with the dynamics of flow. Flow \citep{Hoffman.1996b} has been theorized as a state of deep engagement that produces a distorted sense of time, thus prolonging the duration of user browsing. Accordingly, our model returns a ``sticky'' state with a longer duration as one would expect when users are in flow (and, analogously, for ``non-flow''). We observe characteristics of flow only during browsing and not in the course of goal-directed behavior. This follows \citet{Novak.2003}, according to whom flow is particularly relevant in playful and exploratory behavior.

We explicitly account for duration dependence in our model. Duration dependence has previously received attention in marketing literature. For instance, a customer's propensity to churn is not constant in time but rather dependent on the duration of being a customer \citep{Fader.2018}. In our work, we introduce duration dependence in the transition dynamics of the HMM-based framework. Specifically, we consider the duration of a latent state, which itself is latent (opposed to the time spent on page, which is observable) and which must thus be modeled as a latent variable. It is crucial to emphasize that both result in vastly different behavior: simply considering the time spent on a page  \citep[\eg, as in][]{Ding.2015,Montgomery.2004} makes some pages more likely depending on fast users click, whereas we consider the duration of a user intent and thus how long users were in a state of goal-directed or browsing behavior. 

Our work contributes a novel, duration-dependent HMM. In doing so, we connect to a growing body of HMMs in marketing literature \citep[\eg,][]{abhishek2012media,Ascarza.2013,Ascarza.2018,Ding.2015,gopalakrishnan2021can,Li.2011,Montgomery.2004,Montoya.2010,Netzer.2008,schwartz2014model,Schweidel.2011,Zhang.2014}. HMMs are typically based on the Markov property, according to which a latent state can only be dependent on the previous latent state but not the previous duration in a latent state. The latter is, however, critical for handling a distorted sense of time, and so, to fill this gap, we develop an HMM with duration dependence in the latent dynamics. The resulting model belongs to the wider class of hidden semi-Markov models. We expect our model to aid marketing researchers in modeling related problems that are driven by duration dependence (\eg, learning effects). 

\subsection{Limitations and Future Research}
\label{sec::limitations}

As with other research, our work is not free of limitations that provide an interesting avenue for future research. First, our model finds two states -- ``sticky'' and ``non-sticky'' browsing -- that share similarities with the concept of flow. However, further research is needed to confirm that these indeed capture flow (\eg, through physiological measurements). Second, our partner company informed us that the vast majority of users are not ``logged in'' (\ie, more than 90\,\%), and, hence, customer variables cannot be obtained for them. Due to this, we followed requests of our partner company and specifically tailored our model to a cold-start setting \citep{padilla2021overcoming}. This is also practical importance due to ongoing privacy regulations (\eg, EU General Data Protection Regulation). Marketers can still use our model in settings where customer variables are available and simply incorporate additional information in the covariates. If the entire customer history is available, we further expect that techniques from deep representation learning may be particularly valuable \citep[\eg,][]{ozyurt2022deep}. Third, we performed a case study where we use curated pages as interventions. This choice was made by our partner company, which informed us that it relies primarily on curated pages as part of their marketing efforts. Nevertheless, further interventions such as coupons and price promotions could be included in a similar manner as prior works \citep{Ding.2015, Montgomery.2004, park2018effects}, that is, as covariates in the model. Moreover, the causal effect of these interventions has to be estimated from observational data \citep[\eg,][]{hatt2021estimating}. This may be achieved by using deconfounding methods \citep[\eg, ][]{kuzmanovic2021deconfounding, hatt2021sequential}. Fourth, our numerical results point toward evidence confirming the effectiveness of marketing pages with curated content as interventions. Nevertheless, we acknowledge that an additional experimental evaluation in the field could be a promising roadmap for future research.

\subsection{Concluding Remarks}

The objective of this paper is to predict which users of e-commerce websites are at risk of exiting without a purchase. By making such predictions, online retailers can dynamically adapt their marketing resources such that customers are steered towards marking a purchase. To help in this task, we developed a tailored hidden Markov model with duration dependence. Based on our mathematical model, marketers can detect user exits without purchase more accurately than with models from prior literature. 

  \newcommand{\dq}{"}
\bibliographystyle{agsm}
{\OneAndAHalfSpacedXI
\bibliography{ref}
}

\DoubleSpacedXI
\begin{appendices}
\begin{center}
\LARGE\bfseries Supplements
\end{center}

\section{Estimation Details}
\label{appendix:estimation_details}

\subsection{MCMC Sampling}

Our implementation draws upon recent advances in Bayesian estimation \citep{Gelman.2013}, namely, the Hamiltonian Monte Carlo technique together with the No-U-Turn sampler (NUTS) from Stan \citep{carpenter2017stan}. This approach (often termed \textquote{fully Bayesian}) differs from other estimation techniques, specifically the Metropolis-Hastings algorithm or maximum likelihood estimation. In contrast to these methods, our approach leverages an explicit derivation of likelihood in order to directly sample from the posterior distribution. This is known to be considerably more efficient and, together with Hamiltonian Monte Carlo, requires fewer chains/iterations by a several orders of magnitude \citep{Gelman.2013}. One advantage of theses methods is that we do not need to derive a maximum likelihood approach, an expectation-maximization algorithm, or a Metropolis-Hastings scheme. Instead, a derivation of the likelihood $\mathcal{L}$ is sufficient (see Appendix~\ref{appendix:derivationlikelihood} below).

For each model, we ran four Markov chains, each with a total of \num[group-minimum-digits=3]{4000} iterations. We discarded the initial \num[group-minimum-digits=3]{1000} iterations as part of a warm-up, yielding a total of \num[group-minimum-digits=3]{12000} samples for each model. Our estimation was validated by following the guidelines in \citet{Gelman.2013}. Specifically, we checked our model design by testing whether we retrieve the parameters from simulated data. All checks yielded the desired outcomes. In our paper, we report 95\,\% credible intervals (CI) based on the posterior mean distribution.

We addressed label switching (\ie, states being ordered differently in each Markov chain) by following the recommendations in \citet{jasra2005markov}. This leaves the findings unchanged as it simply orders the states according to their stickiness ($c_{1}\leq \ldots \leq c_{K}$) and thus ensures model identifiability across different runs.

\subsection{Derivation of Likelihood}
\label{appendix:derivationlikelihood}

We derive the likelihood function of the complete \model model using the so-called forward algorithm \citep{Rabiner.1989} as follows. Recall that user $i$ spends $T_i$ periods on the website. The probability of observing a page sequence $o_{i, 1:T_i} = o_{i1},\dots, o_{iT_i}$ is given by the sum over all possible combinations of latent states, duration of these latent states, and duration of the previous latent state. Hence, the probability for observing a page sequence combines the probability of all parameters in the \model. We denote the parameters in the \model by $\lambda_i = \{(Q_i^d)_{d\geq 1}, p_{it\mid s}^{o_{it}}, \theta, c,\pi\}$. Based on this, we write the likelihood $\mathcal{L}(o_{i, 1:T_i}\mid \lambda_i)$ for observing a page sequence via 
\begin{equation*}
\mathcal{L}(o_{i, 1:T_i}\mid \lambda_i)=\sum_{s'\in\mathcal{S}}\sum_{d\geq 1}\alpha_{T_i}(s',d'),
\end{equation*}
where $\alpha_t(s',d')$ are the forward probabilities which are recursively defined via
\begin{equation}\label{eq:alpha}
\begin{aligned}
\alpha_t(s',d') = \sum_{s\in\mathcal{S}\setminus{\{s'\}}}\sum_{d\geq 1} \alpha_{t-d'}(s,d)& \, P(S_{i, t-d'+1}=s'\mid S_{i, t-d'}=s, D_{is}=d) \, P(D_{is'}=d') \\
&\times\prod_{\tau=t-d+1}^{t} \, P(O_{i\tau} = o_{i\tau}\mid S_{i, \tau}=s).
\end{aligned}
\end{equation}
We can rewrite forward probabilities as
\begin{equation*}
\alpha_t(s',d') = \sum_{s\in\mathcal{S}\setminus{\{s'\}}}\sum_{d\geq 1} \alpha_{t-d'}(s,d)q_{iss'}^{d} \, P(D_{is'}=d'\mid \theta_{s'}, c_{s'}) \prod_{\tau=t-d'+1}^{t} p_{i\tau\mid s'}^{o_{i\tau}},
\end{equation*}
using the notation from \Cref{eq:transitionprob,eq:weibullsojourn,eq:emission}.

There are two major differences between the original forward algorithm (used to estimate HMMs) and our modified forward algorithm (used to estimated our \model). First, there are no self-transitions, \ie, once a user decided to leave a state, the user will transition to another, different state. This impacts the forward algorithm with regards to the integration over all possible states the user can transition to. In particular, while the original forward algorithm integrates over all states, our modified forward algorithm only integrates over the states which are different from the current state. Second, the duration in the current state (\ie, the entire history) has to be taken into account. For this, we integrate over all possible durations in the current state. This adds an additional sum (over all possible durations) to the original forward algorithm. Adding this sum increases the computational complexity of our forward algorithm compared to the original forward algorithm. However, we can upper bound the maximum duration in the current state by recognizing that the maximum duration in the current state is upper bounded by the number of pages visited in the current session.

\subsection{Weakly Informative Priors}
\label{appendix:priors}

We choose weakly informative priors \citep{Gelman.2013} for all model parameters. (i)~The initial state distribution $\pi$ of the states is given by $\pi_s = P(S_{i1} = s)$ for $s \in \mathcal{S}$. We choose $\pi\sim\textup{Dir}(1,\ldots,1)$, which is equivalent to a uniform distribution over the open $(K - 1)$-simplex. (ii)~We choose a zero mean normal distribution with a standard deviation of $5$ for the transition parameters $\mu_{ss'}$ and $\delta_{ss'}$. (iii)~The state duration parameters $\theta_s$ and $c_s$ inside the discrete Weibull distribution are set to a standard uniform distribution $\mathcal{U}([0,\ 1])$ for $\theta_s$ and to a normal distribution with mean and standard deviation of $1$ for the parameter $c_s$. The discrete Weibull distribution reduces to a memory-less geometric distribution when $c_s =1$. Hence, the likeliest state duration distribution is the geometric distribution. (iv)~We place zero mean normal distributions with a standard deviation of $5$ on the emission parameters $\gamma_{iss'}$ and $\beta_{iss'}$.

\section{State Duration of the HMM in \citet{Montgomery.2004}}
\label{appendix:distribution_latent_state_durations}

Prior HMMs such as the commonly used HMM in \citet{Montgomery.2004} have a state duration that follows a geometric distribution, and, hence, the expected duration for each state is one. This is shown in the following.

Let $D_{is}$ be a random variable describing the duration in state $s$. Following \citet[p.\,165]{Zucchini.2016}, we can show that for the above HMM, the distribution of its duration $D_{is}$ of state $s$ is given by
\begin{equation}
P\left(D_{is} = d\right) = a_{iss}^{d-1}(1-a_{iss}).
\end{equation}
Hence, the state duration of an HMM follows a geometric distribution which implies that the most likely duration for every state of an HMM is one for $a_{iss} < 1$. Using Bayes's rule and the geometrically distributed state duration, it is straightforward to show that the renewal probability is given by
\begin{equation}\label{eq:geomprobremaining2}
\rho_s(d) = \frac{P\left(D_{is} \geq d+1\right)}{P\left(D_{is} \geq d\right)} = \frac{a_{iss}^{d}}{a_{iss}^{d-1}} = a_{iss}.
\end{equation}
In other words, the probability of remaining in the current state is $a_{iss}$ regardless of the duration in the current state. As a consequence, a user's decision of whether to renew the current state is entirely independent of its duration $d$.

In contrast, the state duration in the \model follows a discrete Weibull distribution. This allows for heterogeneity in the state durations. Likewise, it also changes the \model so that the renewal probability of a state is no longer constant but rather dependent on the previous duration in that state.

\section{Performance Comparison when Accounting for Unobserved Heterogeneity in Transition Mechanism}
\label{appx:sec:DDHMM_UH}

In order to study the effect of unobserved heterogeneity in the transition mechanism of \model, we allow the parameter $\mu_{iss'}$ to vary across individuals. Similar to \citet{netzer2017hidden} and \citet{Ascarza.2018}, this is achieved by modeling $\mu_{iss'} = \phi_{ss'} + \eta_{iss'}$, where $\phi_{ss'}$ represents the average propensity and $\eta_{iss'}$ represents the individual variation.

In \Cref{tbl:model_comparison_DDHMM_UH}, we compare our proposed \model to the above variant, which accounts for unobserved heterogeneity (denoted as \model + UH). Both models use the exact same specification, including the same number of states. We observe that, although the variant \model + UH achieves better results on the training set, our proposed \model achieves superior results on the test set. This is due to the fact that the performance of \model + UH substantially drops on the test set. This is potentially due to overfitting the data, since the \model + UH model introduces an individual parameter for each session, which enlarges the parameter space substantially.

\begin{table}
	\SingleSpacedXI
	\footnotesize
	\centering
	\captionof{table}{Performance Comparison when Accounting for Unobserved Heterogeneity.\label{tbl:model_comparison_DDHMM_UH}}
	\begin{tabular}{lc SSS SSS}  
		\toprule
		&& \multicolumn{3}{c}{In-sample (training data)} &\multicolumn{3}{c}{Out-of-sample (test set)} \\
		\cmidrule(r){3-5}
		\cmidrule(r){6-8}
		Model & \#States & {AUROC} & {AUPRC} & {Hit rate} & {AUROC} & {AUPRC} & {Hit rate}\\
		\midrule
		\model + UH & 3 & 0.7486& 0.1674 & 0.6941 & 0.6155 & 0.1070 & 0.4156 \\
		\model & 3 & 0.7033& 0.1191 & 0.4886 & 0.7176 & 0.1364 & 0.5844 \\
		\bottomrule
		\multicolumn{8}{p{14cm}}{\emph{Note:} \model is our proposed model. \model + UH is the same as \model, but, in addition, accounts for unobserved heterogeneity in the transition mechanism. Higher values are better.}
	\end{tabular}
\end{table}

\section{State-Dependent Behavior Parameter Estimates}
\label{appx:sec:params}

\Cref{tbl:Emission_covariates} reports the posterior means for the coefficients $\gamma_{is}^o$ and $\beta_{is}^o$. Recall from \Cref{sec:model} that we allow $\gamma_{is}^o$ and $\beta_{is}^o$ to depend on the user's latent state. Hence, we receive three potentially different parameters for each $o\in\mathcal{O}$.

As we have seen in \Cref{sec:charlatentstates}, users predominantly view and buy products during {goal-directed} behavior. We see from the table that the visit depth (the number of clicks since the beginning of the session) has a mitigating effect on the probability of viewing ($-2.01$). Simultaneously, the larger the visit depth, the more likely we are to observe a purchase. Hence, users usually start by viewing items and after having collected enough information, they become more likely to make a purchase. Interestingly, the time span (time spent viewing the last item) has the opposite effect. The longer users spend on an item, the more likely they are to view another item again (and the less likely the are to buy an item).

Users in ``sticky'' browsing are mostly characterized by non-directed and impulsive behavior. Therefore, on average, there is no clear tendency of viewing, buying, or exiting; all three types of behavior are possible. However, we see that visit depth mitigates the probability of exiting ($-2.40$) in ``sticky'' browsing, whereas the same covariate reinforces an exiting behavior ($0.69$) in the ``non-sticky'' browsing state. Although the time spent viewing an item (time span) reinforces the probability of viewing in both browsing states, the effect differs. A longer time spent viewing items is linked to a higher likelihood of exiting without a purchase becomes in during ``sticky'' browsing. However, the opposite is true for ``non-sticky'' browsing: a longer time spent viewing an item makes it more likely that a purchase will take place.

\begin{table}
\SingleSpacedXI
\centering
\tiny
\begin{tabular}{lccccccc} 
\toprule
& $\gamma_{is}^o$ & Visit depth & Time span & Cum. pages visited & Weekend\\
\midrule
\csname @@input\endcsname Emission_covariates 
\bottomrule
\end{tabular}
\captionof{table}{Averages of the Individual Posterior Means of $\gamma_{is}^o$ and $\beta_{is}^o$}
\label{tbl:Emission_covariates}
\end{table}

\section{HMM Parameter Estimates}

This section reports detailed estimation results for all components of the HMM in \citet{Montgomery.2004} from \Cref{sec:results}. In the following, we refer to this as ``MLSL model''. Recall that the MLSL model is without duration dependence. The estimates are provided for the variant using $K = 2$ states as this choice is favored during the above model selection.

\subsection{Characterization of Latent States.}
\label{sec:charlatentstates_hmm}

The HMM returns two latent states with different characteristics. In line with \citet{Ding.2015} and \cite{Montgomery.2004}, we label the states as: (i)~{goal-directed} and (ii)~{browsing}. The rationale for this naming is as follows:
\begin{itemize}
	\item {{Goal-directed}:} This state is characterized by an elevated probability of purchase. When in this state, users spend a lot of time on \textsc{Product} pages. Similar to \cite{Montgomery.2004}, this state captures goal-directed behavior (\eg, focused navigation or a purchase orientation).
	\item {{Browsing}:} In this state, user behavior shows a tendency towards collecting information. This is reflected by a high probability of visiting \textsc{Overview} and \textsc{Product} pages. However, the risk of exits is non-zero in this state. This state captures the user's tendency to browse (\eg, participating in a non-purchase orientation).
\end{itemize}

\subsection{Emission Probabilities.}
\label{sec:statedepbehavior_hmm}

Similar to the \model, the emission probabilities of the HMM describe the state-dependent behavior, \ie, the probability of observing a page $O_{it}$ conditional on a latent state. \Cref{tbl:Emission_average_hmm} displays the posterior means of the HMM with average covariates. The probabilities are obtained by averaging over all covariates inside the logit model. In the goal-directed state, users are more likely to make a purchase, but also more likely to exit the website without a purchase. Moreover, users frequently visit \textsc{Product} pages as they have a clear goal in mind. In the browsing state, users predominantly visit \textsc{Account} and \textsc{Product} pages that foster experiential search. Based on these page views, users can collect information on product offerings and product details, respectively. The estimated probability of purchases (\ie, \textsc{Checkout} or \textsc{Order}) in this state is close to zero. Although exits are less likely in the browsing state than in the goal-directed state, the probability of exits is estimated to be non-zero. 

\begin{table}
\SingleSpacedXI
\footnotesize
\centering
\begin{tabular}{lccccccccccc}  
\toprule
& \textsc{Home} & \textsc{Account} & \textsc{Overview} & \textsc{Prod.} & \textsc{Marketing} & \textsc{Comm.}& \textsc{Checkout} & \textsc{Order} & \textsc{Exit}  \\
\midrule
\csname @@input\endcsname Emission_average_hmm 
\bottomrule
\end{tabular}
\captionof{table}{Posterior Means of the State-Dependent Behavior with Average Covariates for HMM.}
\label{tbl:Emission_average_hmm}
\end{table}

\subsection{Distribution of Latent State Durations.}

\Cref{tbl:sojourn_time_coef_hmm} reports the expected duration of latent states in the HMM. The {goal-directed} state possesses a long expected state duration. The estimated duration amounts to $\mathbb{E}[D_s] = 40.24$ time periods. In contrast, a short latent state duration is observed for the {browsing} state, where the duration amounts to $\mathbb{E}[D_s] = 4.11$ time periods.

\begin{table}
	\SingleSpacedXI
	\footnotesize
	\centering
	\begin{tabular}{lccc}  
		\toprule
		&$\theta_s$ &$\mathbb{E}[D_s]$ \\
		\midrule
    \csname @@input\endcsname duration_parameters_hmm 
		\bottomrule
	\end{tabular}
	\captionof{table}{Averages of the Individual Posterior Means of the state duration coefficients for HMM.}
	\label{tbl:sojourn_time_coef_hmm}
\end{table}

\begin{figure}
\SingleSpacedXI
    \centering
\begin{tikzpicture}[x=1pt,y=1pt]
\definecolor{fillColor}{RGB}{255,255,255}
\path[use as bounding box,fill=fillColor,fill opacity=0.00] (0,0) rectangle (361.35,180.67);
\begin{scope}
\path[clip] (  0.00,  0.00) rectangle (361.35,180.67);
\definecolor{drawColor}{RGB}{255,255,255}
\definecolor{fillColor}{RGB}{255,255,255}

\path[draw=drawColor,line width= 0.6pt,line join=round,line cap=round,fill=fillColor] (  0.00, -0.00) rectangle (361.35,180.67);
\end{scope}
\begin{scope}
\path[clip] ( 31.64, 25.92) rectangle (190.99,160.72);
\definecolor{fillColor}{RGB}{255,255,255}

\path[fill=fillColor] ( 31.64, 25.92) rectangle (190.99,160.72);
\definecolor{drawColor}{gray}{0.92}

\path[draw=drawColor,line width= 0.3pt,line join=round] ( 31.64, 47.36) --
	(190.99, 47.36);

\path[draw=drawColor,line width= 0.3pt,line join=round] ( 31.64, 78.00) --
	(190.99, 78.00);

\path[draw=drawColor,line width= 0.3pt,line join=round] ( 31.64,108.64) --
	(190.99,108.64);

\path[draw=drawColor,line width= 0.3pt,line join=round] ( 31.64,139.28) --
	(190.99,139.28);

\path[draw=drawColor,line width= 0.3pt,line join=round] ( 63.03, 25.92) --
	( 63.03,160.72);

\path[draw=drawColor,line width= 0.3pt,line join=round] (111.32, 25.92) --
	(111.32,160.72);

\path[draw=drawColor,line width= 0.3pt,line join=round] (159.61, 25.92) --
	(159.61,160.72);

\path[draw=drawColor,line width= 0.6pt,line join=round] ( 31.64, 32.04) --
	(190.99, 32.04);

\path[draw=drawColor,line width= 0.6pt,line join=round] ( 31.64, 62.68) --
	(190.99, 62.68);

\path[draw=drawColor,line width= 0.6pt,line join=round] ( 31.64, 93.32) --
	(190.99, 93.32);

\path[draw=drawColor,line width= 0.6pt,line join=round] ( 31.64,123.96) --
	(190.99,123.96);

\path[draw=drawColor,line width= 0.6pt,line join=round] ( 31.64,154.60) --
	(190.99,154.60);

\path[draw=drawColor,line width= 0.6pt,line join=round] ( 38.88, 25.92) --
	( 38.88,160.72);

\path[draw=drawColor,line width= 0.6pt,line join=round] ( 87.17, 25.92) --
	( 87.17,160.72);

\path[draw=drawColor,line width= 0.6pt,line join=round] (135.46, 25.92) --
	(135.46,160.72);

\path[draw=drawColor,line width= 0.6pt,line join=round] (183.75, 25.92) --
	(183.75,160.72);
\definecolor{drawColor}{RGB}{0,0,128}

\path[draw=drawColor,line width= 1.7pt,line join=round] ( 38.88,151.55) --
	( 54.98,151.55) --
	( 71.08,151.55) --
	( 87.17,151.55) --
	(103.27,151.55) --
	(119.37,151.55) --
	(135.46,151.55) --
	(151.56,151.55) --
	(167.66,151.55) --
	(183.75,151.55);
\definecolor{drawColor}{gray}{0.20}

\path[draw=drawColor,line width= 0.6pt,line join=round,line cap=round] ( 31.64, 25.92) rectangle (190.99,160.72);
\end{scope}
\begin{scope}
\path[clip] (196.49, 25.92) rectangle (355.85,160.72);
\definecolor{fillColor}{RGB}{255,255,255}

\path[fill=fillColor] (196.49, 25.92) rectangle (355.85,160.72);
\definecolor{drawColor}{gray}{0.92}

\path[draw=drawColor,line width= 0.3pt,line join=round] (196.49, 47.36) --
	(355.85, 47.36);

\path[draw=drawColor,line width= 0.3pt,line join=round] (196.49, 78.00) --
	(355.85, 78.00);

\path[draw=drawColor,line width= 0.3pt,line join=round] (196.49,108.64) --
	(355.85,108.64);

\path[draw=drawColor,line width= 0.3pt,line join=round] (196.49,139.28) --
	(355.85,139.28);

\path[draw=drawColor,line width= 0.3pt,line join=round] (227.88, 25.92) --
	(227.88,160.72);

\path[draw=drawColor,line width= 0.3pt,line join=round] (276.17, 25.92) --
	(276.17,160.72);

\path[draw=drawColor,line width= 0.3pt,line join=round] (324.46, 25.92) --
	(324.46,160.72);

\path[draw=drawColor,line width= 0.6pt,line join=round] (196.49, 32.04) --
	(355.85, 32.04);

\path[draw=drawColor,line width= 0.6pt,line join=round] (196.49, 62.68) --
	(355.85, 62.68);

\path[draw=drawColor,line width= 0.6pt,line join=round] (196.49, 93.32) --
	(355.85, 93.32);

\path[draw=drawColor,line width= 0.6pt,line join=round] (196.49,123.96) --
	(355.85,123.96);

\path[draw=drawColor,line width= 0.6pt,line join=round] (196.49,154.60) --
	(355.85,154.60);

\path[draw=drawColor,line width= 0.6pt,line join=round] (203.74, 25.92) --
	(203.74,160.72);

\path[draw=drawColor,line width= 0.6pt,line join=round] (252.03, 25.92) --
	(252.03,160.72);

\path[draw=drawColor,line width= 0.6pt,line join=round] (300.32, 25.92) --
	(300.32,160.72);

\path[draw=drawColor,line width= 0.6pt,line join=round] (348.61, 25.92) --
	(348.61,160.72);
\definecolor{drawColor}{RGB}{0,0,128}

\path[draw=drawColor,line width= 1.7pt,line join=round] (203.74,124.81) --
	(219.83,124.81) --
	(235.93,124.81) --
	(252.03,124.81) --
	(268.12,124.81) --
	(284.22,124.81) --
	(300.32,124.81) --
	(316.41,124.81) --
	(332.51,124.81) --
	(348.61,124.81);
\definecolor{drawColor}{gray}{0.20}

\path[draw=drawColor,line width= 0.6pt,line join=round,line cap=round] (196.49, 25.92) rectangle (355.85,160.72);
\end{scope}
\begin{scope}
\path[clip] ( 31.64,160.72) rectangle (190.99,175.17);
\definecolor{drawColor}{gray}{0.20}
\definecolor{fillColor}{gray}{0.85}

\path[draw=drawColor,line width= 0.6pt,line join=round,line cap=round,fill=fillColor] ( 31.64,160.72) rectangle (190.99,175.17);
\definecolor{drawColor}{gray}{0.10}

\node[text=drawColor,anchor=base,inner sep=0pt, outer sep=0pt, scale=  0.64] at (111.32,165.75) {Goal-Directed};
\end{scope}
\begin{scope}
\path[clip] (196.49,160.72) rectangle (355.85,175.17);
\definecolor{drawColor}{gray}{0.20}
\definecolor{fillColor}{gray}{0.85}

\path[draw=drawColor,line width= 0.6pt,line join=round,line cap=round,fill=fillColor] (196.49,160.72) rectangle (355.85,175.17);
\definecolor{drawColor}{gray}{0.10}

\node[text=drawColor,anchor=base,inner sep=0pt, outer sep=0pt, scale=  0.64] at (276.17,165.75) {Browsing};
\end{scope}
\begin{scope}
\path[clip] (  0.00,  0.00) rectangle (361.35,180.67);
\definecolor{drawColor}{gray}{0.20}

\path[draw=drawColor,line width= 0.6pt,line join=round] ( 38.88, 23.17) --
	( 38.88, 25.92);

\path[draw=drawColor,line width= 0.6pt,line join=round] ( 87.17, 23.17) --
	( 87.17, 25.92);

\path[draw=drawColor,line width= 0.6pt,line join=round] (135.46, 23.17) --
	(135.46, 25.92);

\path[draw=drawColor,line width= 0.6pt,line join=round] (183.75, 23.17) --
	(183.75, 25.92);
\end{scope}
\begin{scope}
\path[clip] (  0.00,  0.00) rectangle (361.35,180.67);
\definecolor{drawColor}{gray}{0.30}

\node[text=drawColor,anchor=base,inner sep=0pt, outer sep=0pt, scale=  0.64] at ( 38.88, 16.56) {1};

\node[text=drawColor,anchor=base,inner sep=0pt, outer sep=0pt, scale=  0.64] at ( 87.17, 16.56) {4};

\node[text=drawColor,anchor=base,inner sep=0pt, outer sep=0pt, scale=  0.64] at (135.46, 16.56) {7};

\node[text=drawColor,anchor=base,inner sep=0pt, outer sep=0pt, scale=  0.64] at (183.75, 16.56) {10};
\end{scope}
\begin{scope}
\path[clip] (  0.00,  0.00) rectangle (361.35,180.67);
\definecolor{drawColor}{gray}{0.20}

\path[draw=drawColor,line width= 0.6pt,line join=round] (203.74, 23.17) --
	(203.74, 25.92);

\path[draw=drawColor,line width= 0.6pt,line join=round] (252.03, 23.17) --
	(252.03, 25.92);

\path[draw=drawColor,line width= 0.6pt,line join=round] (300.32, 23.17) --
	(300.32, 25.92);

\path[draw=drawColor,line width= 0.6pt,line join=round] (348.61, 23.17) --
	(348.61, 25.92);
\end{scope}
\begin{scope}
\path[clip] (  0.00,  0.00) rectangle (361.35,180.67);
\definecolor{drawColor}{gray}{0.30}

\node[text=drawColor,anchor=base,inner sep=0pt, outer sep=0pt, scale=  0.64] at (203.74, 16.56) {1};

\node[text=drawColor,anchor=base,inner sep=0pt, outer sep=0pt, scale=  0.64] at (252.03, 16.56) {4};

\node[text=drawColor,anchor=base,inner sep=0pt, outer sep=0pt, scale=  0.64] at (300.32, 16.56) {7};

\node[text=drawColor,anchor=base,inner sep=0pt, outer sep=0pt, scale=  0.64] at (348.61, 16.56) {10};
\end{scope}
\begin{scope}
\path[clip] (  0.00,  0.00) rectangle (361.35,180.67);
\definecolor{drawColor}{gray}{0.30}

\node[text=drawColor,anchor=base east,inner sep=0pt, outer sep=0pt, scale=  0.64] at ( 26.69, 29.84) {0.00};

\node[text=drawColor,anchor=base east,inner sep=0pt, outer sep=0pt, scale=  0.64] at ( 26.69, 60.48) {0.25};

\node[text=drawColor,anchor=base east,inner sep=0pt, outer sep=0pt, scale=  0.64] at ( 26.69, 91.12) {0.50};

\node[text=drawColor,anchor=base east,inner sep=0pt, outer sep=0pt, scale=  0.64] at ( 26.69,121.75) {0.75};

\node[text=drawColor,anchor=base east,inner sep=0pt, outer sep=0pt, scale=  0.64] at ( 26.69,152.39) {1.00};
\end{scope}
\begin{scope}
\path[clip] (  0.00,  0.00) rectangle (361.35,180.67);
\definecolor{drawColor}{gray}{0.20}

\path[draw=drawColor,line width= 0.6pt,line join=round] ( 28.89, 32.04) --
	( 31.64, 32.04);

\path[draw=drawColor,line width= 0.6pt,line join=round] ( 28.89, 62.68) --
	( 31.64, 62.68);

\path[draw=drawColor,line width= 0.6pt,line join=round] ( 28.89, 93.32) --
	( 31.64, 93.32);

\path[draw=drawColor,line width= 0.6pt,line join=round] ( 28.89,123.96) --
	( 31.64,123.96);

\path[draw=drawColor,line width= 0.6pt,line join=round] ( 28.89,154.60) --
	( 31.64,154.60);
\end{scope}
\begin{scope}
\path[clip] (  0.00,  0.00) rectangle (361.35,180.67);
\definecolor{drawColor}{RGB}{0,0,0}

\node[text=drawColor,anchor=base,inner sep=0pt, outer sep=0pt, scale=  0.80] at (193.74,  7.06) {Latent state duration $d$};
\end{scope}
\begin{scope}
\path[clip] (  0.00,  0.00) rectangle (361.35,180.67);
\definecolor{drawColor}{RGB}{0,0,0}

\node[text=drawColor,rotate= 90.00,anchor=base,inner sep=0pt, outer sep=0pt, scale=  0.80] at ( 11.01, 93.32) {Renewal Probability $\rho_s(d)$};
\end{scope}
\end{tikzpicture}
	\caption{Renewal probabilities in the states (i)~goal-directed search and (ii)~{browsing} as a function of the latent state duration $d$ for HMM.}
\label{fig:renewalprobability_hmm}
\end{figure}
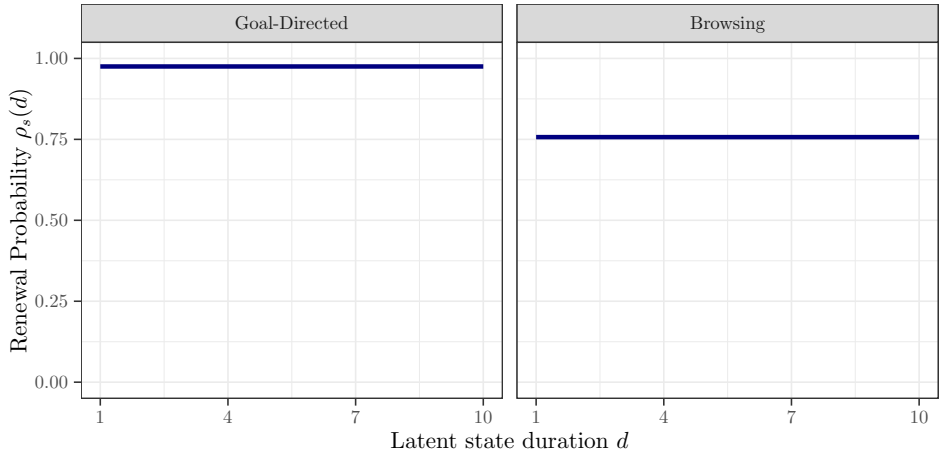

The traditional HMM has no duration dependence. Nevertheless, we can compute the renewal probability similar to our \model, which allows us to make a detailed comparison. See \Cref{sec:theorymodel} for a derivation of the renewal probability. Note that, due to the absence of duration dependence, the renewal probability in the traditional HMM is a constant function. The corresponding renewal probability is illustrated in \Cref{fig:renewalprobability_hmm}. The main ramification here is that the renewal probability remains constant in time for the HMM. Hence, the renewal probability is duration-independent, since the HMM cannot capture such dependencies.

\subsection{Initial Latent State Distribution.}

The initial distribution over latent states for the HMM, $\pi$, describes the probability of a user starting her web session in a certain state. The probability estimates are displayed in \Cref{tbl:Initial probability transition_hmm}. Evidently, most users (52\,\%) start their web session in a {browsing} state. Hence, we find that the predominant share of users starts their web sessions with experiential activities and thus begin by collecting information. In comparison, 48\,\% of all users start their web session in a {goal-directed} state. 

\begin{table}
\SingleSpacedXI
\footnotesize
\centering
\begin{tabular}{lcc}  
\toprule
    & Posterior mean  &\\
\midrule
\csname @@input\endcsname initial_hmm 
\bottomrule
\end{tabular}
\captionof{table}{Posterior Means of Belonging to Each of the States at the Beginning of each Session for HMM.}
\label{tbl:Initial probability transition_hmm}
\end{table}

\section{Simulation Study: Parameter Identification}
\label{apx:simulation_study}

In this section, we describe the simulation study, which validates that our estimation procedure identifies the true set of the parameters of the \model model. We refer to the parameters via the set $\lambda_i = \{(Q_i^d)_{d\geq 1}, p_{it\mid s}^{o}, \theta, c,\pi\}$, where $(Q_i^d)_{d\geq 1}$ describes the duration-dependent transition probabilities, $p_{it\mid s}^{o}$ denotes the emission probabilities for each state and each page, $\theta$ and $c$ determine the distribution of the state duration, and $\pi$ is the initial state distribution. We then use the following data-generating process. 

The observed pages are sampled from a logit model similar to \Cref{eq:emission}, \ie,
\begin{equation}
	p_{it\mid s}^o = \frac{\e^{\gamma_{is}^o + \beta_{is}^o x_{it}}}{\sum\limits_{k \in \mathcal{O}} \e^{\gamma_{is}^k + \beta_{is}^k x_{it}}},
\end{equation}
where $\gamma_{is}^o$ and $\beta_{is}^o$ are sampled from $\mathcal{N}(0, 5^2)$. Furthermore, the duration distribution is given similarly to \Cref{eq:weibullsojourn} as
\begin{equation}
	P\left(D_{is} = d\mid \theta_s, c_s\right) = (1-\theta_s)^{(d-1)^{c_s}} - (1-\theta_s)^{d^{c_s}} \quad \text{ for }0<\theta_s<1,\ c_s>0,
\end{equation}
where $c_s$ and $\theta_s$ are sampled from $\mathcal{N}(0, 1)$ and $\mathcal{U}([0, 1])$, respectively. Moreover, the transition probability is, similarly to \Cref{eq:transitionprob}, sampled from 
\begin{equation}
	q_{iss'}^d = \frac{e^{\mu_{ss'} + \delta_{ss'}d}}{\sum_{k\in\mathcal{S}\setminus \{s\}}{e^{\mu_{sk} + \delta_{sk}d}}},
\end{equation}
where $\mu_{ss'}$ and $\delta_{ss'}$ are samples from $\mathcal{N}(0, 5^2)$.  Finally, the initial state probabilities, $\pi$, are sampled from Dir$(1, 1, 1)$.

We sample from this data-generating process for a number of sequences (\ie, sessions) ranging in $\{50, 100, 200, 300, 400, 500\}$. We average the results over 10 runs over the data-generating process, where we use a maximum a posteriori estimate in every run. The results (\ie, the error compared to the true parameter) are presented for the parameters $\gamma$, $\pi$, $c_s$, $\theta$, $\mu$, $\delta$ in \Cref{fig:sim_param_1} and for the parameter $\beta$ in \Cref{fig:sim_param_2}. In both figures, we observe that, for increasing numbers of sequences (\ie, sessions), the estimates parameters converge to the true parameters. Hence, our estimation procedure correctly identifies the true parameters of the data-generating process.

\begin{turn}{-90}
\begin{minipage}{\linewidth}
\begin{figure}
\SingleSpacedXI
    \centering
    \scalebox{0.35}{\includegraphics{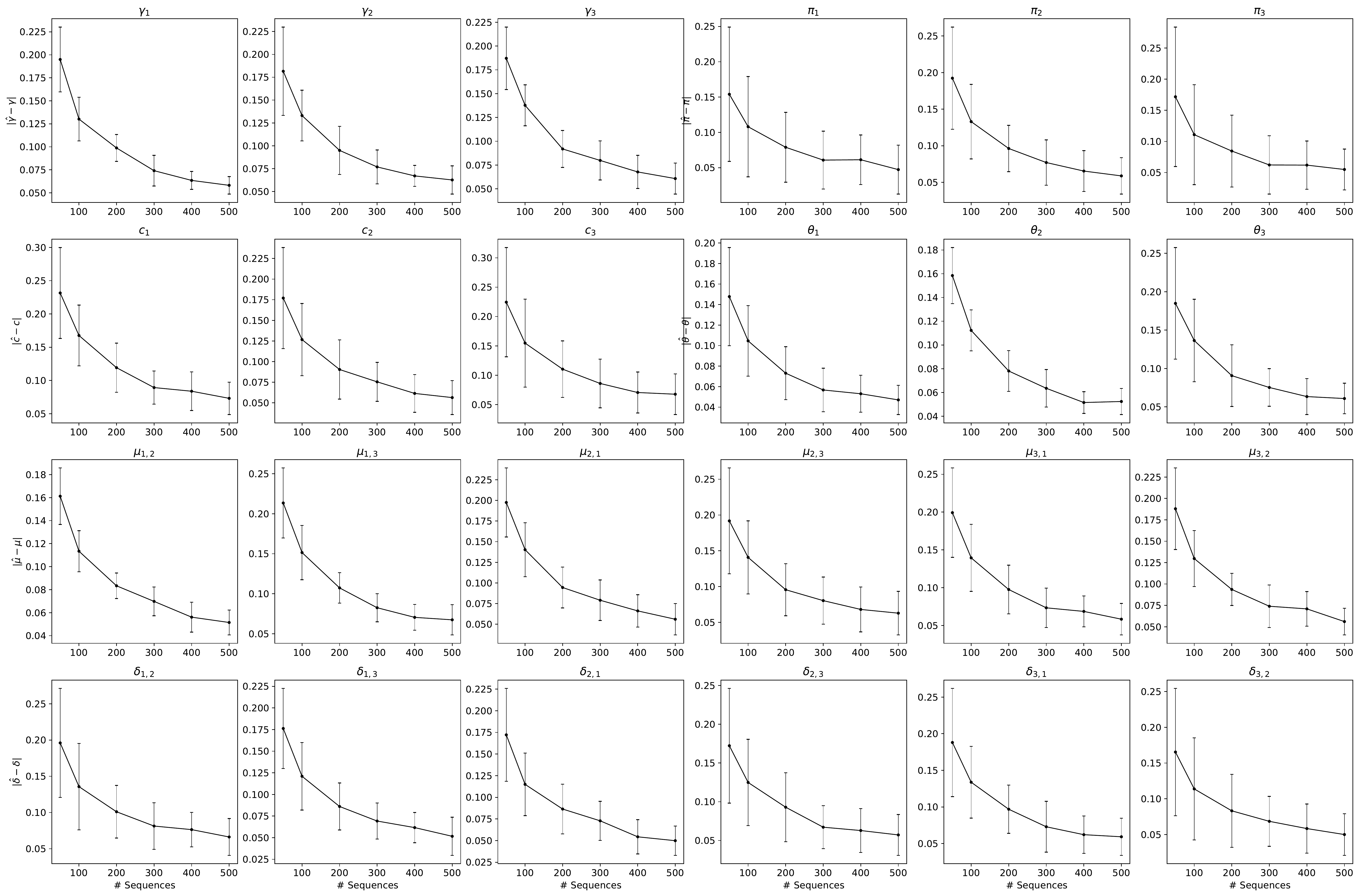}}
	\parbox{21cm}{\caption{Results of simulation study: Difference between estimated parameter, $\hat{\gamma}, \hat{\pi}, \hat{c}_s, \hat{\theta}, \hat{\mu}, \hat{\delta}$, and true parameter, $\gamma, \pi, c_s, \theta, \mu, \delta$, for different number of sequences (\ie, sessions). The mean and standard deviation are given over 10 runs. We observe that the error decays the more sequences we use in our estimation procedure. Hence, the true parameters are correctly estimated.}}
\label{fig:sim_param_1}
\end{figure}
\end{minipage}
\end{turn}

\begin{turn}{-90}
\begin{minipage}{\linewidth}
\begin{figure}
\SingleSpacedXI
    \centering
    \scalebox{0.275}{\includegraphics{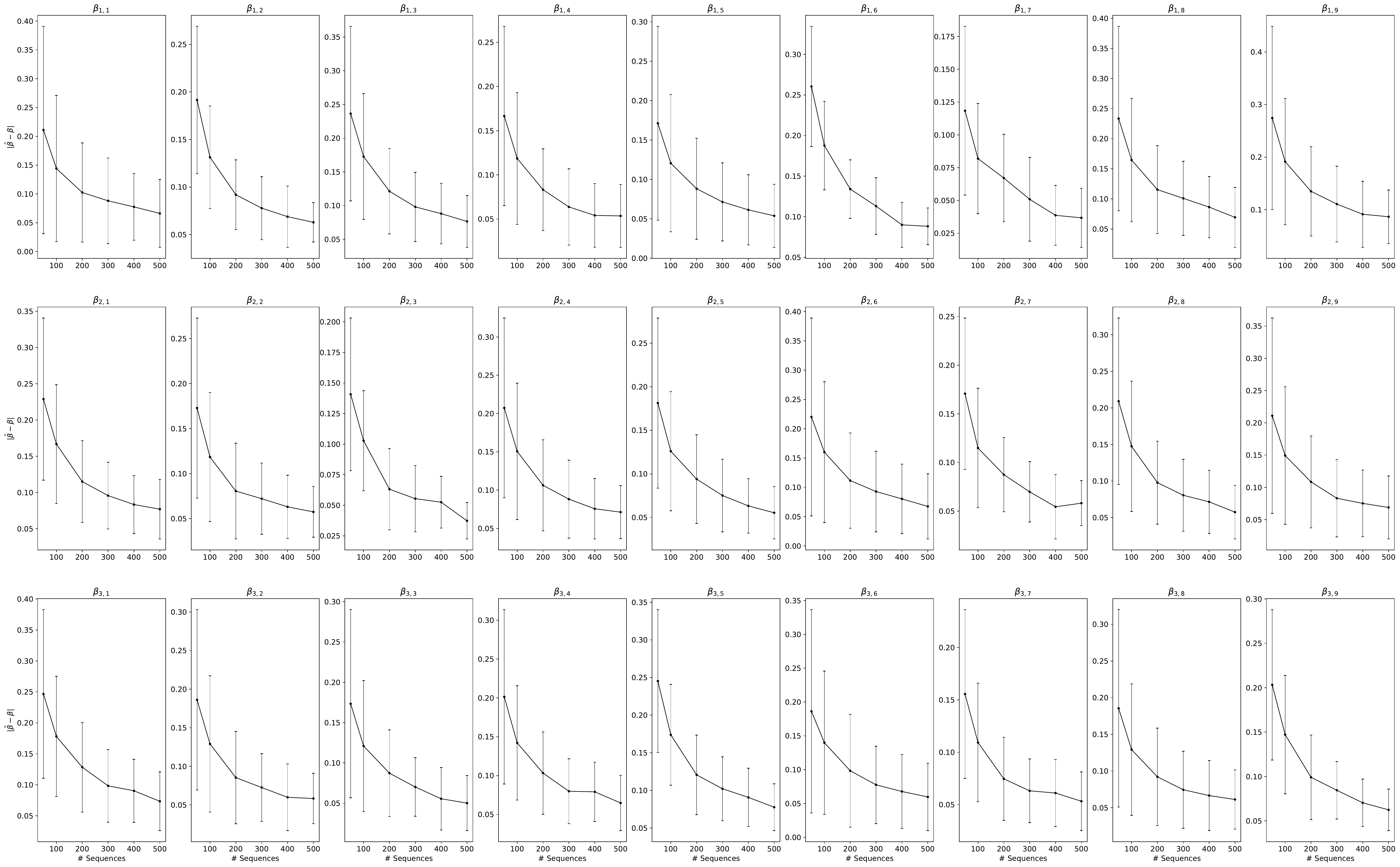}}
	\parbox{21cm}{\caption{Results of simulation study: Difference between estimated parameter, $\hat{\beta}$, and true parameter, $\beta$, for different number of sequences (\ie, sessions). The mean and standard deviation are given over 10 runs. We observe that the error decays the more sequences we use in our estimation procedure.  Hence, the true parameters are correctly estimated.}}
    \label{fig:sim_param_2}
\end{figure}
\end{minipage}
\end{turn}

\end{appendices}

\end{document}